\documentclass[journal]{IEEEtran}
\usepackage[utf8]{inputenc}
\usepackage{multicol}
\usepackage{multirow}
\usepackage[keeplastbox]{flushend}
\usepackage{diagbox}
\usepackage{amsmath}
\usepackage{booktabs}
\usepackage{multirow}
\usepackage{verbatim}
\usepackage{array}
\usepackage{dblfloatfix}
\usepackage{fixltx2e}
\usepackage{algorithmic}
\usepackage{algorithm}
\usepackage{array}
\usepackage{url}
\usepackage{float}
\usepackage{graphicx,psfrag,epsfig}
\usepackage{caption}
\usepackage{subfigure}
\usepackage{color}
\usepackage[export]{adjustbox}
\usepackage{cite}
\usepackage{xspace}

\begin{document}
\title{Oil Spill SAR Image Segmentation via Probability Distribution Modelling}
\author{Fang Chen, Aihua Zhang, Heiko Balzter, Peng Ren and Huiyu Zhou
\thanks{Fang Chen and Huiyu Zhou are with School of Computing and Mathematical Sciences, University of Leicester, Leicester LE1 7RH, United Kingdom. E-mail: fc160@leicester.ac.uk; E-mail: hz143@leicester.ac.uk. (Corresponding author: Huiyu Zhou.)}
\thanks{Aihua Zhang is with Division of Science and Technology, Beijing Normal University-Hongkong Baptist University UIC, Zhuhai 519087, China.
E-mail: evaahzhang@uic.edu.cn.}
\thanks{Heiko Balzter is with School of Geography, University of Leicester, Leicester LE1 7RH, United Kingdom. E-mail: hb91@leicester.ac.uk.}
\thanks{Peng Ren is with the College of Oceanography and Space Informatics, China University of Petroleum (East China), Qingdao 266580, China. E-mail: pengren@upc.edu.cn.}}


\maketitle

\begin{abstract}
Segmentation of marine oil spills in Synthetic Aperture Radar (SAR) images is a challenging task because of the complexity and irregularities in SAR images. In this work, we aim to develop an effective segmentation method which addresses marine oil spill identification in SAR images by investigating the distribution representation of SAR images. To seek effective oil spill segmentation, we revisit the SAR imaging mechanism in order to attain the probability distribution representation of oil spill SAR images, in which the characteristics of SAR images are properly modelled. We then exploit the distribution representation to formulate the segmentation energy functional, by which oil spill characteristics are incorporated to guide oil spill segmentation. Moreover, the oil spill segmentation model contains the oil spill contour regularisation term and the updated level set regularisation term which enhance the representational power of the segmentation energy functional. Benefiting from the synchronisation of SAR image representation and oil spill segmentation, our proposed method establishes an effective oil spill segmentation framework. Experimental evaluations demonstrate the effectiveness of our proposed segmentation framework for different types of marine oil spill SAR image segmentation.
\end{abstract}

\section{introduction}
\IEEEPARstart{M}{arine} oil spills, which are the release of liquid petroleum hydrocarbons into the marine environmental areas, occur at different scales and lead disastrous consequences to the natural marine ecosystem. These disasters have posed challenges to marine environmental protection \cite{kingston2002long} and are difficult to be cleaned up. In this regard, detecting oil spills efficiently has become extremely important in the field of geoscience and remote sensing \cite{brekke2005oil}. Satellite-based Synthetic Aperture Radar (SAR) works as a powerful tool for remotely monitoring environmental and natural targets \cite{soukissian2016satellite} \cite{li2017wind} on the earth surface by virtue of its all-weather observation ability and independence of day and night conditions. Additionally, because of its high spatial resolution and temporal repetition frequency, SAR has been an important tool for monitoring marine oil spills \cite{fiscella2000oil} \cite{solberg2007oil}. Particularly, in actual operations for marine environmental protection, to conduct timely oil spill damage assessment and control the spread of oil spills, it is vital to accurately detect oil spill regions in SAR images. Therefore, detection of marine oil spills in SAR images has been an important research topic in the marine remote sensing community \cite{topouzelis2008oil}. In this paper, we concentrate on accurately detecting marine oil spills in SAR images by developing an intelligent method which considers SAR image formation and oil spill segmentation simultaneously.

In exploring the relationship between SAR image formation and oil spill segmentation, SAR based marine oil spill monitoring presents seminal image data for image segmentation \cite{gemme2018automatic}, and monitoring with different sensors enables thorough exploration of the observed maritime targets \cite{velotto2016first}. In the remote sensing literature, oil spills on the ocean surface have different scattering properties at microwaves from the surrounding unpolluted waters \cite{ulaby2014microwave} \cite{alpers2017oil}. Therefore, we take a close look at the non-Bragg scattering phenomenon caused by oil spills, which is extremely important for characterising oil spills in a SAR image. Specifically, the capillary and short gravity waves on the ocean surface give rise to Bragg scattering that is sensed by SAR \cite{yin2014extended}. In contrast, oil spills dampen or suppress the generation of capillary and short gravity waves and thus result in dark patches in SAR images where non-Bragg scattering dominates. In this scenario, the non-Bragg scattering is responsible for interpreting oil spills in the observations.

From physical meaning perspective, SAR polarimetry is an important research topic in oil spill observation \cite{migliaccio2007sar}. In this regard, more and more oil spill polarimetric characteristics have been explored recently, and state-of-the-art studies on oil spill detection include those by Migliaccio et al. \cite{nunziata2014sea}, Ricci et al. \cite{bandiera2005slicks} \cite{bandiera2013bayesian}, Minchew et al. \cite{minchew2012polarimetric} \cite{collins2015use} and Brekke et al. \cite{brekke2016cross} \cite{espeseth2017analysis}. These strategies take advantages of the physical properties of oil spills for oil spill detection in SAR images \cite{salberg2014oil}. Additionally, the analysis of the polarimetric properties of oil spills has helped to formulate techniques such as thresholding \cite{lupidi2017fast} and k-means clustering \cite{buono2016polarimetric} to perform effective oil spill detection \cite{liu2017coastline}.

In the image processing perspective, recent advances on oil spill segmentation have drawn much attention of researchers who have concentrated working on developing sophisticated strategies for segmenting oil spill areas in SAR images. Generally, most of these strategies are formulated based on energy minimisation \cite{zhang2015level} \cite{jing2011novel}, where a segmentation energy functional is developed with respect to fitness and similarity to measure the characteristic homogeneity of image pixels. For example, Li et al. \cite{li2010distance} exploited a reinitialisation free technique to formulate the segmentation energy functional for avoiding numeric remedy. Yin et al. \cite{yin2014modified} developed a multiphase segmentation technique by introducing a piecewise constant model which properly localises region boundaries. Xia et al. \cite{xia2015meaningful} employed the standard level set method to develop a modified continuous energy functional with multi-scale and non-local characterisation for segmenting oil spill regions. Mdakane et al. \cite{mdakane2017image} presented a continuous level set based energy functional to detect oil spills from moving vessels with the incorporation of region-based signed pressure force. Chen et al. \cite{chen2017level} developed a segmentation energy functional by incorporating the self-guided filtering scheme for edge preserving. Ren et al. \cite{ren2018energy} proposed a one dot fuzzy initialisation strategy for efficient marine oil spill segmentation. Li et al. \cite{li2021oil} developed a multiscale conditional adversarial network which trains an oil spill detection model with small data. Furthermore, to address the challenges of blurry SAR image segmentation, Chen et al. \cite{chen2018segmenting} developed a blurry image segmentation method with the exploitation of the alternating direction method of multipliers (ADMM), which effectively detects oil spill regions in blurry SAR images. These strategies are normally developed from the image processing perspective alone, and do not explicitly consider mining the formation of oil spill images which characterises the internal characteristics of oil spills. Therefore, the segmentation accuracy of these techniques is heavily depending on the visual appearance of objects in SAR images.

In addition, most of the state-of-the-art studies treat the oil spill image generating and image processing as two fully independent stages, e.g. \cite{espeseth2017analysis} \cite{yu2018oil}. However, these two procedures are correlated internally, where the segmentation performance heavily depends on the features generated from the images.

To achieve efficient and accurate oil spill segmentation, depending on the prior knowledge, in this paper, we present a novel oil spill SAR image segmentation framework in which both SAR image formation and oil spill segmentation are considered simultaneously. Specifically, to construct an effective segmentation framework, we commence by investigating SAR imaging for marine oil spills, by which the internal characteristics of oil spills are modelled. We then formulate the segmentation energy functional with the underpinning distribution model which can guide the segmentation operation to conduct an optimal oil spill segmentation. Particularly, in SAR imaging for marine oil spills, the SAR images are generated after the radar receives the backscatters from the ocean surface. Thus, to obtain SAR image formation, it is critical to explore the intermediate transformations, and the procedures responsible for generating SAR image formation are presented in section \ref{SAR observation of oil spills}. We then construct the segmentation energy functional in which the image distribution model and the segmentation indicator are incorporated. Benefiting from the oil spill distribution model, the proposed segmentation method conducts accurate and effective oil spill segmentation.

In summary, the main contributions of our work include:

$\bullet$ We propose a novel oil spill SAR image segmentation method, which incorporates SAR image formation and oil spill segmentation simultaneously. The proposed segmentation framework exploits the inherent properties of SAR images to guide effective segmentation.

$\bullet$ We formulate the objective function by constructing the oil spill segmentation energy functional with fitness measure, contour regularisation and update regularisation. This results in a tractable optimisation scheme for segmenting irregular shaped oil spills in SAR images.

$\bullet$ We develop an efficient segmentation algorithm that converges to optimal contours using the standard gradient descent method to solve the formulated segmentation energy functional. This enables accurate numerical computations for oil spill contours and thus achieves accurate segmentation.

Experimental evaluations for different types of marine oil spill SAR image segmentation are conducted to validate the effectiveness of our proposed segmentation method over both visual and quantitative metrics.

\section{SAR Imaging for Marine Oil Spills}
\label{SAR observation of oil spills}
SAR imaging is an important technique to generate satisfactory SAR image data and thus investigating SAR imaging for marine oil spills enables us to obtain the representation of oil spill SAR images. In this section, we investigate the formation of SAR images.

\subsection{Probability Modelling of Marine Oil Spill SAR Images}
\label{exponential distribution for oil spill SAR image description}
In SAR imaging, the presence of oil spills on the ocean surface alters the radar scattering properties of the ocean, resulting in the backscatters different from those where oil spills are not present \cite{ulaby2014microwave}. This manifests that the received backscatters from the oil spill areas are different from those of the other areas. Specifically, the relation between the observed targets and the backscattered power to the radar receiver can be depicted by the radar equation. According to \cite{ulaby2014microwave}, we have the equation of the SAR imaging radar shown as follows:
\begin{equation}
\begin{split}
 \langle P^{r}(\theta) \rangle&= [ \frac{P^{t}\lambda^{2}G^{2}(\theta)lc\eta }{4(4\pi)^{3}R^{4}\sin\theta } ]\sigma(\theta )\\
 &=K\sigma(\theta )
 \end{split}
 \label{SAR imaging radar equation}
\end{equation}
where $P^{r}$ is the power delivered to the radar receiver, $P^{t}$ is the power from the transmitter, $G$ is the gain of the antenna, $\theta$ is the incidence angle, $R$ is the distance between the radar and the illuminated area, $\eta$ is the length of narrow pulse, $l$ is the along-track length of the antenna, $\sigma$ is the radar cross section (RCS) component and $K$ represents the quantity in the square bracket. Eq. (\ref{SAR imaging radar equation}) shows that the viewing angle is included in the antenna gain and $\sin$. This indicates that these two parameters depend on $\theta$. Additionally, in practice, the sensor-target distance $R$ also depends on $\theta$. For oil spill SAR image segmentation, we aim to obtain the representation of SAR image formation to formulate the segmentation model. According to \cite{ulaby2014microwave}, the representation of SAR image $I$ is correlated to the detection manner of the received signal, and according to \cite{li2010oil} \cite{oliver2004understanding}, we have the distribution representation of SAR image $I$ in terms of the observed intensity at each pixel shown as follows:
\begin{equation}
\begin{split}
p(I)&=\frac{1}{\bar{I}}{\rm exp}^{(-\frac{1}{\bar{I}}I)}\\
&=\frac{1}{K_{s}\sigma}{\rm exp}^{(-\frac{1}{K_{s}\sigma}I)}, \sigma>0
\end{split}
\label{SAR image intensity density function}
\end{equation}
where $K_{s}$ is the detection system constant. This representation shows that the probability distribution of the SAR image $I$ is subject to the exponential distribution. In practice, exponential, Weibull and Gamma distributions are often exploited to describe SAR imagery \cite{oliver2004understanding} \cite{gao2019characterization}.

\subsection{Distribution Models of SAR imagery}
In the previous sub-section, we introduced the exponential distribution model. For fair comparisons, we further discuss two representative distribution models that may be exploited in analysing SAR imagery in the following parts.

\subsubsection{Weibull distribution for SAR image characterisation}
The Weibull distribution has been recognised as one of the commonly used models to describe SAR imagery, according to \cite{oliver2004understanding}, which is of the intensity as follows:
\begin{equation}
p(I)=\frac{\upsilon}{\sigma ^{\upsilon }}I^{\upsilon -1}{\rm exp}^{[-(\frac{I}{\sigma})^{\upsilon }]}, \upsilon>0
\end{equation}
where $\upsilon$ is the shape parameter which controls the shape of the distribution. Specifically, the Weibull distribution can be used to precisely describe speckle noise, whilst characterising high-resolution SAR images.

\subsubsection{Gamma distribution for SAR image characterisation}
Using the characterisation of SAR imagery \cite{oliver2004understanding}, we derive the Gamma distribution as follows:
\begin{equation}
 p(I)=\frac{1}{\Gamma(\kappa )}(\frac{\kappa}{\sigma})^{\kappa}I^{\kappa-1}{\rm exp}^{(-\frac{\kappa I}{\sigma })}, \kappa>0
\end{equation}
where $\Gamma(\kappa)$ is Gamma function, and $\kappa$ is the order parameter which together with the RCS component $\sigma$ characterise the Gamma probability distribution. The Gamma distribution is widely used to characterise speckle noise. In addition, the Gamma distribution is more applicable for describing terrains such as the grassland regions in high-resolution SAR images.

\subsection{Remarks}
The Weibull and Gamma distributions are widely used for describing SAR imagery.
Oil spills on the ocean surface dampen the surface waves by decreasing the surface tension. This results in a smoother surface that reduces the surface reflectivity than the surrounding water areas, and therefore oil spill patches appear dark with lower intensities. As pointed out in \cite{oliver2004understanding}, the Weibull distribution generates an incorrect probability for lower intensities. Moreover, the Weibull distribution is more applicable in high-resolution SAR images, however, maritime scenes are of low-resolution images. Furthermore, the Gamma distribution is effective to describe general noise process or speck noise. Thus, it is not recommended to use the Weibull or Gamma distribution to describe oil spill SAR images.
In \cite{oliver2004understanding}, the information of each image pixel is carried by the RCS component $\sigma$. Thus, $\sigma$ is the information-bearing parameter, which can be estimated from the SAR image distribution model.

\section{Energy Functional for Marine Oil Spill Segmentation}
\label{oil spill segmentation formulation}
\begin{figure*}[!htbp]	
	\centering
	\includegraphics[width=1.02\textwidth,height=0.33\textheight]{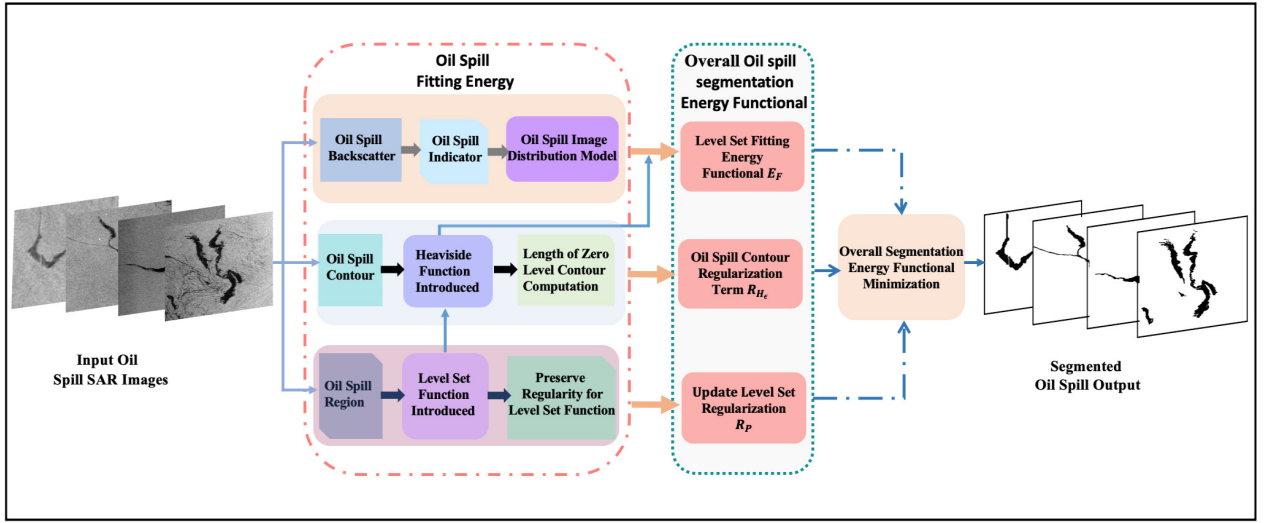}
	\vspace*{-7mm}
	\caption[]{The framework of the proposed oil spill SAR image segmentation method. This segmentation framework consists of two main modules (i.e., the oil spill fitting energy and the overall oil spill segmentation energy functional). Specifically, for the oil spill fitting energy module, it consists of three components: the energy fitness component which is constructed with the exploitation of the image probability distribution model, the oil spill contour outlining component which is formulated by using the Heaviside function on the basis of zero level sets, and the level set regularisation component which is introduced to ensure stable evolution. The second module also consists of three components, and each of which corresponds to its counterpart in the first module. The components in this module are then integrated to construct the overall segmentation energy functional. Minimization of the overall segmentation energy functional achieves effective oil spill segmentation.}
	\label{our method flowchart}	
\end{figure*}

Segmentation of oil spill SAR images is a very challenging task in practice, because oil spills in SAR images exhibit irregular shapes. This demonstrates that exploration the formation of oil spill SAR images is helpful for effective oil spill segmentation. In this paper, we develop a novel oil spill segmentation method by simultaneously considering SAR imaging and oil spill segmentation. Specifically, our proposed method exploits the probability distribution function of oil spill SAR images to construct the segmentation framework. Thus, the segmentation is inclined to implement accurate oil spill segmentation. A brief demonstration of our proposed method is illustrated in Fig. \ref{our method flowchart}. This figure shows that the segmentation framework consists of two main parts: the oil spill fitting energy part in which the image distribution model is incorporated to effectively extract oil spill areas, the length of the zero level contour term which is introduced to approximate the length of the oil spill contours and the level set regularisation term which is designed for accurate computation and stable level set evolution. The other part is the overall oil spill segmentation energy functional, which is corresponding to the oil spill fitting energy counterpart. Details for constructing the segmentation energy functional are explained in the following subsections.

\subsection{Probability Distribution for Constructing Energy Functional}
\label{pdf formulation segmentation energy functional}
In image segmentation, the segmentation energy functional describes the object of interest based on the integrated elements. For oil spill SAR image segmentation, the segmentation objective is to accurately identify the oil spill areas. Specifically, given an oil spill SAR image $I$, the image domain is categorised as the oil spill and non-oil spill areas, and the whole image domain $\Omega$ can be represented as $\Omega=\cup _{i=1}^{2}\Omega _{i}$ and $\Omega _{1}\cap \Omega _{2}=\O$. Particularly, the intensity distributions of the oil spill and non-oil spill areas are different, where each area is modelled with one distribution. To give a specific representation for the image intensity, for a given point $\textbf x$ in the image domain, we define a neighbouring area centered $\textbf x$ as $O_{\textbf x}$. According to Eq. (\ref{SAR image intensity density function}), we have the distribution corresponding to domain $\Omega_{i}$ shown as follows:
\begin{equation}
p(I(\textbf y)\mid\sigma_{i},\textbf{x})=\frac{1}{K_{s}\sigma_{i}(\textbf x)}{\rm exp}^{\Big(-\frac{1}{K_{s}\sigma_{i}(\textbf x)}I(\textbf y)\Big)}
\label{SAR image intensity density function in domain}
\end{equation}
which describes dynamic regions in a local image domain. $I(\textbf y)$ are the intensities involved in the local region and $\sigma_{i}$ is the component that characterises the distribution. To generalise Eq. (\ref{SAR image intensity density function in domain}), we have the following relation:
\begin{equation}
 p\Big(I(\textbf x)\mid\sigma _{i}\Big)\propto \prod_{\textbf y\in \Omega _{i}\cap O_{\textbf x}}p\Big(I(\textbf y)\mid\sigma _{i}, \textbf x\Big)
\end{equation}
This relation shows that in region $O_{\textbf x}$, the distribution of image intensity $I(\textbf x)$ is correlated to samples $\big\{I(\textbf y), \textbf y\in \Omega_{i}\cap O_{\textbf x} \big\}$.

In an oil spill SAR image, oil spills are randomly distributed in the image domain. Thus, for accurate oil spill image segmentation, the segmentation procedure is expected to operate across the entire image domain. To achieve this, we describe the image intensity distribution from the entire image domain perspective. Referring to \cite{solberg2007oil}, the intensities of image pixels are independently distributed, and with \cite{bishop2006pattern}, let $S$ = $\{ I(\textbf x)\mid \sigma_{i}$, $\textbf x \in \Omega$, i=1,2\}, we have the following likelihood function:
\begin{equation}
p(S\mid \sigma_{i})=\prod_{\textbf x\in \Omega}p\Big(I(\textbf x)\mid\sigma _{i}\Big)
\end{equation}
and the joint distribution function is given as follows:
\begin{equation}
\begin{split}
p(S\mid \sigma)&=\prod_{i=1}^{2}p(S\mid \sigma_{i})=\prod_{i=1}^{2}\prod_{\textbf x\in\Omega}p\Big(I(\textbf x)\mid\sigma_{i}\Big)\\
&\propto\prod_{i=1}^{2}\prod_{\textbf x\in\Omega}\prod_{\textbf y\in \Omega _{i}\cap O_{\textbf x}}p\Big(I(\textbf y)\mid\sigma _{i}, \textbf x\Big)
\end{split}
\label{oil spill likelihood function}
\end{equation}
This function describes the distribution from an entire image perspective. In the operation of oil spill SAR image segmentation, the segmentation energy functional is constructed with the exploitation of the image distribution representation. Following the relationships between the segmentation energy functional and the joint distribution likelihood function \cite{goodfellow2016deep} \cite{zhang2015level}, we construct the segmentation energy functional that is the inverse log-likelihood function of $p(S\mid \sigma)$ in Eq. (\ref{oil spill likelihood function}) across the entire image domain:
\begin{equation}
\begin{split}
E_{\sigma}&=-\int_{\Omega}\log p(S\mid\sigma)d\textbf x\\
&={\rm constant}-\!\sum_{i=1}^{2}\int_{\Omega}\!\int_{\Omega_{i}\cap O_{\textbf x}}\!\!{ \log}\bigg(p\big(I(\textbf y)\mid\sigma_{i},\textbf x\big)\bigg)d\textbf yd\textbf x
\end{split}
\label{energy function with constant}
\end{equation}
This energy functional operates fitness with respect to image intensities. The log-likelihood in the energy functional shows that $\sigma$ plays an important role in indicating the segmentation fitness and the double integral defines the segmentation operation across the entire image domain. In oil spill SAR image segmentation, to constrain the segmentation fitting energy functional, a kernel function $K(\textbf f)$, which is capable of adapting the intensity $I(\textbf y)$ involved in local region $O_{\textbf x}$, is introduced to determine the scalable ranges during the segmentation evolution \cite{mercier2006partially}. According to \cite{li2008minimization}, the kernel function with a scale parameter $\tau>0$ is chosen as follows:
\begin{equation}
K_{\tau}(\textbf f)=\frac{1}{\big(\sqrt{2\pi}\tau\big)^{2}}{\rm exp}^{\big(-\frac{|\textbf f|^{2}}{2\tau ^{2}}\big)}
\label{region indicator function}
\end{equation}
In practice, the kernel function $K_{\tau}(\textbf f)$ also benefits effective oil spill segmentation.

Specifically, for centre point $\textbf x$, when its neighbouring point $\textbf y$ deviates, the kernel function uses the detailed form of $K_{\tau}(\textbf x,\textbf y)=\frac{1}{(\sqrt{2\pi}\tau)^{2}}{\rm exp}^{(-\frac{|\textbf x-\textbf y|^{2}}{2\tau ^{2}})}$ to measure the away distance and adapt the neighbouring region scale. Putting the distribution in Eq. (\ref{SAR image intensity density function in domain}) and the kernel function $K_{\tau}(\textbf x,\textbf y)$ into Eq. (\ref{energy function with constant}) and eliminating the trivial constant terms, the oil spill SAR image segmentation energy functional can be rewritten as follows:
\begin{equation}
\begin{split}
E\Big(\sigma_{i}(\textbf x)\Big)\!\!=\!\!\sum_{i=1}^{2}\!\!\int_{\Omega}\!\!\int_{\Omega}\!\! K_{\tau}(\textbf x,\textbf y)\Big[\frac{ I(\textbf y)}{K_{s}\sigma_{i}(\textbf x)}+{\rm ln}\Big(\sigma_{i}(\textbf x)\Big)\Big]d\textbf yd\textbf x
\end{split}
\label{the energy function without constant}
\end{equation}
The integration used in this functional illustrates that the segmentation computation is applied to the whole image domain.

\subsection{Level Set Segmentation Fitting Energy Functional}
\label{level set fitting term}
In oil spill SAR image segmentation, to manifest the performance of the segmentation operation, a segmentation indicator of quantifying the segmentation progress is required. Level set methods, which have the advantage to perform numerical computations describing curves and surfaces on a fixed Cartesian grid without having to parameterise these objects \cite{li2017level}. Moreover, a level set method is capable of tracing shapes that change their topology in an efficient way. Thus, in the implementation of oil spill SAR image segmentation, we exploit the level set method to perform accurate computations during curves evolving to search for the object oil spill regions. Specifically, in level set methods, a contour $C\subset \Omega$ is represented by the zero level set of a Lipschitz function $\phi:\Omega\rightarrow\Re$, namely the zero level set function. In this scenario, we utilise level set function $\phi$ to represent the oil spill image segmentation and outside and inside the contour $C$, the level set function $\phi$ takes positive and negative values respectively. We formulate the segmentation energy functional with level set function as follows:
\begin{equation}
\begin{split}
E_{F}\Big(\sigma_{i}(\textbf x),\phi(\textbf y)\!\Big)\!=&\gamma_{1}\!\!\int_{\Omega}\!\!\int_{\Omega} \!\!\!\!K_{\tau}(\textbf x,\textbf y)\!\Big[\frac{ I(\textbf y)}{K_{s}\sigma_{1}(\textbf x)}\!+\!{\rm ln}\Big(\sigma_{1}(\textbf x)\Big)\Big]\\
&H_{\epsilon}\Big(\phi(\textbf y)\Big)d\textbf yd\textbf x +\gamma_{2}\int_{\Omega}\int_{\Omega} K_{\tau}(\textbf x,\textbf y)\\
&\!\Big[\!\frac{ I(\textbf y)}{K_{s}\sigma_{2}(\textbf x)}
\!+\!{\rm ln}\Big(\!\sigma_{2}(\textbf x)\!\Big)\!\Big]\!\Big(\!1\!-\!H_{\epsilon}\big(\phi(\textbf y)\!\big)\!\Big)\!d\textbf yd\textbf x\\
=&\!\sum_{i=1}^{2}\!\!\gamma_{i}\!\!\int_{\Omega}\!\!\int_{\Omega}\!\!\!\! K_{\tau}(\textbf x,\textbf y)\!\Big[\!\frac{I(\textbf y)}{K_{s}\sigma_{i}(\textbf x)}\!+\!\ln\!\Big(\sigma_{i}(\textbf x)\!\Big)\!\Big]\\
&M_{i}^{\epsilon}\Big(\phi(\textbf y)\Big)d\textbf yd\textbf x
\end{split}
\label{level set energy functional for fitness}
\end{equation}
where $\gamma_{1}$, $\gamma_{2}$ are positive balancing constants. $M_{1}^{\epsilon}(\phi)=H_{\epsilon}(\phi)$, $M_{2}^{\epsilon}(\phi)=1-H_{\epsilon}(\phi)$, and $H_{\epsilon}(\phi)$ is the Heaviside function given as follows:
\begin{equation}
H_{\epsilon}\Big(\phi(\textbf y)\Big)=\frac{1}{2}\Big[1+\frac{2}{\pi}{\rm arctan}(\frac{\phi(\textbf y)}{\epsilon})\Big]
\end{equation}
where $\epsilon$ is a positive smooth parameter introduced to make $H(\phi)$ smooth.

\subsection{Oil Spill Contour Regularisation Term}
\label{oil contour relularity term}
In level set methods, the curve obtained in terms of the intersection between the level set function $\phi$ and the zero plane \big(i.e., \big($\phi(\textbf x)=0\big)$\big) is referred to as the zero-level set of $\phi$. For oil spill SAR image segmentation, the zero-level set of $\phi$ indicates the contours of the oil spill regions \cite{chen2018segmenting}. To compute the length of the zero level set contour of $\phi$, a contour regularisation term is introduced as follows:
\begin{equation}
R_{H_{\epsilon}}\Big(\phi(\textbf x)\Big)=\int\Big|\bigtriangledown H_{\epsilon }\big(\phi(\textbf x)\big)\Big|d\textbf x
\label{contour regularization term}
\end{equation}
This regularisation term is commonly used in variational level set based segmentation methods for regularising the zero level set contour.

\subsection{Update Level Set Regularisation}
\label{update level set regularization}
Oil spills in a SAR image are normally of irregular shapes. Thus, for oil spill SAR image segmentation, the iterative evolution is expected to search for an optimal solution of level set function $\phi$. This can be achieved by updating the level set evolution in a regularisation way. To maintain the regularisation of the level set evolution for accurate computation and stable evolution, we here introduce a distance regularisation term into our variational level set segmentation energy functional \cite{li2008minimization}. This regularisation term is defined with a potential function shown as follows:
\begin{equation}
 R_{P}\big(\phi(\textbf x)\big)=\int\frac{1}{2}\Big(\big| \bigtriangledown \phi(\textbf x)\big|-1\Big)^{2}d\textbf x
 \label{regularization term}
\end{equation}
 The regularisation term $R_{p}(\phi)$ characterises the deviation of the level set function $\phi$ from a signed distance function and also improves the performance of oil spill segmentation.

\subsection {Overall Energy Functional for Oil Spill Segmentation}
\label{overall oil spill segmentation energy functional}
 Segmenting oil spills from SAR images accurately is a systematic work, which is relevant to the reciprocal process introduced in \ref{level set fitting term}, \ref{oil contour relularity term} and \ref{update level set regularization}. Specifically, our proposed segmentation energy functional is formulated with a level set energy term measuring the fitness of oil spills (Eq. (\ref{level set energy functional for fitness})), a regularisation term with respect to oil spill contour outlining (Eq. (\ref{contour regularization term})), and an update regularisation term for accurate computation and stable level set evolution (Eq. (\ref{regularization term})). These terms are integrated to formulate the overall segmentation energy functional, and the integration formulation is given as follows:
 \begin{equation}
 \begin{split}
 E_{S}\big(\sigma_{i},\phi\big)=&E_{F}\big(\sigma_{i}(\textbf x),\phi(\textbf y)\big)+\nu R_{H_{\epsilon}}\big(\phi(\textbf x)\big)\\
 &+\mu R_{P}\big(\phi(\textbf x)\big)
 \label{the final oil spill segmentation energy functional}
 \end{split}
\end{equation}
where $\nu$ and $\mu$ are positive balancing parameters.

\section{Energy Functional Minimisation for Oil Spill Segmentation}
In the proposed oil spill segmentation framework, energy functional minimisation is exploited to seek optimal segmentation, and the procedures of conducting energy functional minimisation are illustrated in the following subsections.

\subsection{Minimisation of the Overall Energy Functional}
To minimise the overall energy functional, we use the following equation:
\begin{equation}
\Big\{\tilde{\sigma _{i}}(\textbf x),\tilde{\phi}(\textbf x)\Big\}={\rm arg\min_{\sigma_{i}(\textbf x),\phi(\textbf x)}}E_{S}\Big(\sigma_{i}(\textbf x),\phi(\textbf x)\Big)
\end{equation}
which indicates the minimisation operation over $\sigma_{i}$ and $\phi$. Particularly, for the representation of the oil spill SAR image, $\sigma_{i}$ is the dominant characteristic of representing oil spills in SAR images. In this scenario, we conduct the minimisation with respect to $\sigma_{i}$ in the first step. This guides the segmentation intentionally towards the oil spill areas. We then conduct the minimisation with respect to the segmentation level set function $\phi$, benefiting from the guidance of $\sigma_{i}$, the segmentation operation anticipates accurate oil spill segmentation. The detailed minimisation computations for these two steps are give in \ref{minimization with respect to oil characteristic and level set function}.

\subsection{Energy Functional Minimisation for Segmenting Oil Spills}
\label{minimization with respect to oil characteristic and level set function}
In energy functional minimisation, the segmentation is updated in an iterative manner, and the alternated minimisation for $\sigma_{i}(\textbf x)$ and $\phi$ in the following two steps leads to effective oil spill segmentation.

In the first step, to obtain $\sigma_{i}(\textbf x)$ for guiding optimal oil spill segmentation, we seek the minimisation of the overall energy functional $E_{S}(\sigma_{i},\phi)$ with respect to $\sigma_{i}(\textbf x)$ whilst the other parameters are fixed, and the equation for minimisation is given as follows:
\begin{equation}
\begin{split}
\int K_{s}\sigma_{i}\big(\textbf x\big)K_{\tau}\big(\textbf x,\textbf y\big)M_{i}^{\epsilon}\Big(\phi(\textbf y)\Big)d\textbf y-\int K_{\tau}\big(\textbf x,\textbf y\big)\\
\Big[M_{i}^{\epsilon}\big(\phi(\textbf y)\big)I(\textbf y)\Big]d\textbf y=0
\end{split}
\label{minimization for oil features}
\end{equation}
From Eq. (\ref{minimization for oil features}), we have the following equivalent equation:
\begin{equation}
\begin{split}
K_{s}\Big[K_{\tau}(\textbf x)*M_{i}^{\epsilon}\big(\phi(\textbf x)\big)\Big]\sigma _{i}(\textbf x)-K_{\tau}(\textbf x)*\\
\Big[M_{i}^{\epsilon}\big(\phi(\textbf x)\big)I(\textbf x)\Big]=0
\end{split}
\label{minimization of oil features}
\end{equation}
where $*$ denotes the convolution operation. According to Eq. (\ref{minimization of oil features}), we obtain the minimisation computation for $\sigma_{i}$ as follows:
\begin{equation}
\sigma_{i}\big(\textbf x\big)=\frac{K_{\tau}(\textbf x)*\Big[M_{i}^{\epsilon }\big(\phi(\textbf x)\big)I(\textbf x)\Big]}{K_{s}\Big[K_{\tau}(\textbf x)*M_{i}^{\epsilon}\big(\phi(\textbf x)\big)\Big]}
\label{minimization of sigma}
\end{equation}
where $K_{\tau}(\textbf x)$ represents the localisation for datapoint $\textbf x$.

In the second step, in order to obtain optimal $\phi$ in iterations, keeping $\sigma_{i}(\textbf x)$ fixed, we compute the minimisation of the energy functional $E_{S}(\sigma_{i},\phi)$ with respect to $\phi$ using the standard gradient descent method to solve the gradient flow equation as follows:
\begin{equation}
\begin{split}
\frac{\partial \phi}{\partial t}=&\big(\gamma_{2}\breve{\varepsilon}_{2}-\gamma_{1}\breve{\varepsilon}_{1}\big)\delta_{\epsilon}\big(\phi\big)+\nu \delta _{\epsilon}\big(\phi\big){\rm div}\Big(\frac{\bigtriangledown \phi}{| \bigtriangledown \phi|}\Big)\\
&+\mu \Big(\bigtriangledown ^{2}\phi-{\rm div}\big(\frac{\bigtriangledown \phi}{|\bigtriangledown \phi |}\big)\Big)
\end{split}
\label{level set contour updating}
\end{equation}
where $\breve{\varepsilon}_{i}, i=1,2$ are introduced to simplify the formulation, which is given as follows:
\begin{equation}
\tilde{\varepsilon _{i}}\big(\textbf x\big)=\!\!\int\!\! K_{\tau}\big(\textbf y,\textbf x\big)\Big[\frac{ I(\textbf x)}{K_{s}\sigma_{i}(\textbf y)}+{\rm ln}\big(\sigma_{i}(\textbf y)\big)\Big]d\textbf y
\label{intermidiate term}
\end{equation}
The function $\delta_{\epsilon}(\phi)$ in (\ref{level set contour updating}) is given as follows:
\begin{equation}
\begin{split}
\delta_{\epsilon} \big(\phi\big)&=\frac{\partial H_{\epsilon}\big(\phi\big)}{\partial \phi}\\
&=\frac{1}{\pi}\frac{\epsilon }{\epsilon ^{2}+\phi^{2}}
\end{split}
\end{equation}

To update the segmentation during the iterative evolution, we alternately update $\sigma_{i}(\textbf x)$ and $\phi$ during the iteration and the system updating from $k$th to $(k+1)$th goes over the following process:
\begin{equation}
\sigma_{i}^{k+1}(\textbf x)=\frac{K_{\tau}(\textbf x)*\Big[M_{i}^{\epsilon }\big(\phi^{k}(\textbf x)\big)I(\textbf x)\Big]}{K_{s}\Big[K_{\tau}(\textbf x)*M_{i}^{\epsilon}\big(\phi^{k}(\textbf x)\big)\Big]}
\label{updating of sigma}
\end{equation}

\begin{equation}
\begin{split}
\phi^{k+1}=&\phi^{k}+t\Big[\big(\gamma_{2}\breve{\varepsilon}_{2}-\gamma_{1}\breve{\varepsilon}_{1}\big)\delta\big(\phi^{k}\big)+\nu \delta _{\epsilon}\big(\phi^{k}\big){\rm div}\big(\frac{\bigtriangledown \phi^{k}}{| \bigtriangledown \phi^{k}|}\big)\\
&+\mu \big(\bigtriangledown ^{2}\phi^{k}-{\rm div}(\frac{\bigtriangledown \phi^{k}}{|\bigtriangledown \phi^{k} |})\big)\Big]
\end{split}
\label{minimization of phi}
\end{equation}
where $k$ is the iteration number. We alternately operate the above two steps till the minimisation of the overall energy functional is reached.

A summation for implementing the proposed oil spill segmentation framework is illustrated in Algorithm \ref{oil spill specific algorithm}. Specifically, we have $I$ as the input oil spill SAR image, and initialise $\phi^0$ by an initialisation box around an oil spill region.

\begin{algorithm}
	\centering
	\caption{Segmenting Oil Spills from SAR Images Based \rightline {on Satellite Imaging for Marine Oil Spill Observation}}
	\label{oil spill specific algorithm}
	\begin{algorithmic}
		\vspace*{4.6pt}
		\STATE\textbf {Input:} Oil spill SAR image $I$.\\	
		\vspace*{5.6pt}
		\STATE  {$K_{s}\leftarrow {\rm the\quad fixed\quad system \quad constant}, \phi ^{0}=\rm initialLSF$}.
		\vspace*{4.6pt}
		\STATE {$K_{\tau }\leftarrow {\rm Gaussian{\quad kernel} }, H_{i}^{\epsilon }\leftarrow {\rm Heaviside\quad function}$}.
		\vspace*{4.6pt}
		\STATE \textbf {For}  from $k=1$ till segmentation convergence: \textbf {do}
		\vspace*{5.6pt}
		\STATE \qquad Solve $\frac{\partial E_{S}(\sigma_{i}, \phi)}{\partial \sigma_{i}}=0$, update $(\sigma_{i})^{k+1}$ based on (\ref{updating of sigma}).
		\STATE \qquad Solve $\frac{\partial E_{S}(\sigma_{i}, \phi)}{\partial \phi}=0$, update $\phi^{k+1}$ based on (\ref{minimization of phi}).
		\vspace*{5.6pt}
		\vspace*{5.6pt}
		\STATE \textbf {End for}
		\vspace*{4.6pt}
		\STATE \textbf {Output:} Segmented oil spill regions in SAR image $I$.\\
		\vspace*{3.6pt}  	
	\end{algorithmic}
\end{algorithm}

\subsection{Remarks}
We present the pipeline of our proposed image segmentation framework and the minimisation of the proposed segmentation energy functional in Sections \ref{oil spill segmentation formulation} and \ref{minimization with respect to oil characteristic and level set function}, respectively. We obtain that the exponential distribution model is applicable for modelling marine oil spill SAR images. To demonstrate the applicability of our proposed algorithm in processing the image data with different distributions, we here compare the object functionals of using our proposed segmentation algorithm under the Weibull and Gamma distributions.

As explained in Section \ref{SAR observation of oil spills}, the Weibull distribution is characterised by the RCS component $\sigma$ and the shape parameter $\upsilon$. Thus, under the Weibull distribution, Eqs. (\ref{the energy function without constant}) and (\ref{level set energy functional for fitness}) in our proposed algorithm are revised as follows:
\begin{equation}
\begin{split}
E_{w}\big(\sigma _{i}(\textbf x)\big)\!\!=\!\!&\sum_{i=1}^{2}\int_{\Omega}\!\!\int_{\Omega}\!\!K_{\tau}(\textbf x, \textbf y)\Big[\upsilon {\rm ln}\big(\sigma_{i}(\textbf x)\big)+\big(\frac{I(\textbf y)}{\sigma_{i}(\textbf x)}\big)^{\upsilon}\\
&-(\upsilon-1){\rm ln}\big(I(\textbf y)\big)\Big]d\textbf yd\textbf x
\end{split}
\end{equation}
which illustrates the revised segmentation fitting energy functional under the Weibull distribution. Similarly, the level set fitting energy functional with the Weibull distribution is given by:
\begin{equation}
\begin{split}
E_{F_{w}}\Big(\sigma_{i}(\textbf x),\phi(\textbf y)\!\Big)\!\!=&\gamma_{1}\!\!\int_{\Omega}\!\!\int_{\Omega} \!\!\!\!K_{\tau}(\textbf x,\textbf y)\!\Big[\upsilon {\rm ln}\big(\sigma_{i}(\textbf x)\big)+\big(\frac{I(\textbf y)}{\sigma_{i}(\textbf x)}\big)^{\upsilon}\\
&-(\upsilon-1){\rm ln}\big(I(\textbf y)\big)\Big]H_{\epsilon}\Big(\phi(\textbf y)\Big)d\textbf yd\textbf x +\gamma_{2}\\
&\int_{\Omega}\!\!\int_{\Omega} K_{\tau}(\textbf x,\textbf y)
\!\Big[\upsilon {\rm ln}\big(\sigma_{i}(\textbf x)\big)+\big(\frac{I(\textbf y)}{\sigma_{i}(\textbf x)}\big)^{\upsilon}\\
&-(\upsilon-1){\rm ln}\big(I(\textbf y)\big)\Big]\!\Big(\!1\!-\!H_{\epsilon}\big(\phi(\textbf y)\!\big)\!\Big)\!d\textbf yd\textbf x\\
=&\!\sum_{i=1}^{2}\!\!\gamma_{i}\!\!\int_{\Omega}\!\!\int_{\Omega}\!\!\!\! K_{\tau}(\textbf x,\textbf y)\!\Big[\!\upsilon {\rm ln}\big(\sigma_{i}(\textbf x)\big)+\big(\frac{I(\textbf y)}{\sigma_{i}(\textbf x)}\big)^{\upsilon}\!\\
&-(\upsilon-1){\rm ln}\big(I(\textbf y)\big)\!\Big]
M_{i}^{\epsilon}\Big(\phi(\textbf y)\Big)d\textbf yd\textbf x
\end{split}
\end{equation}
As shown above, besides Eqs. (\ref{the energy function without constant}) and (\ref{level set energy functional for fitness}), the equations (\ref{minimization for oil features}) and (\ref{intermidiate term}) are also correlated to the distribution representation. Thus, we derive the adapted forms of Eqs. (\ref{minimization for oil features}) and (\ref{intermidiate term}) with respect to the Weibull distribution as follows:
\begin{equation}
\begin{split}
\int \sigma_{i}^{\upsilon}\big(\textbf x\big)K_{\tau}\big(\textbf x,\textbf y\big)M_{i}^{\epsilon}\Big(\phi(\textbf y)\Big)d\textbf y-\int K_{\tau}\big(\textbf x,\textbf y\big)\\
\Big[M_{i}^{\epsilon}\big(\phi(\textbf y)\big)I^{\upsilon}(\textbf y)\Big]d\textbf y=0
\end{split}
\end{equation}
Eq. (\ref{intermidiate term}) is introduced to simplify the representation of the minimisation of $\phi$, which is revised as:
\begin{equation}
\widetilde{\varepsilon _{w_{i}}}\big(\textbf x\big)\!\!=\!\!\int\!\! K_{\tau}\big(\textbf y,\textbf x\big)\!\Big[\!\upsilon{\rm ln}\big(\sigma_{i}(\textbf y)\big)+\big(\frac{ I(\textbf x)}{\sigma_{i}(\textbf y)}\big)^{\upsilon}\!\!-(\upsilon-1){\rm ln}\big(I(\textbf x)\big)\!\Big]d\textbf y
\end{equation}
For the Gamma distribution, Eq. (\ref{the energy function without constant}) can be revised under the Gamma distribution as follows:
\begin{equation}
\begin{split}
E_{g}\big(\sigma _{i}(\textbf x)\big)\!=\!&\sum_{i=1}^{2}\int_{\Omega}\!\!\int_{\Omega}\!\!K_{\tau}(\textbf x, \textbf y)\Big[(\kappa \!-\!1){\rm ln}\big(I(\textbf y)\big)\!-\! \kappa {\rm ln}\big(\sigma_{i}(\textbf x)\big)\\
&-\big(\frac{\kappa I(\textbf y)}{\sigma_{i}(\textbf x)}\big)
-{\rm ln}\Gamma (k)\Big]d\textbf yd\textbf x
\end{split}
\end{equation}
We similarly derive the level set fitting energy functional under the Gamma distribution as follows:
\begin{equation}
\begin{split}
E_{F_{g}}\!\Big(\sigma_{i}(\textbf x),\phi(\textbf y)\!\Big)\!\!&=\!\gamma_{1}\!\!\int_{\Omega}\!\!\int_{\Omega} \!\!\!\!K_{\tau}(\textbf x,\textbf y)\!\Big[\!(\kappa\!-\!\!1){\rm ln}\big(I(\textbf y)\big)\!\!-\! \kappa {\rm ln}\big(\sigma_{i}(\textbf x)\big)\\
&-\big(\frac{\kappa I(\textbf y)}{\sigma_{i}(\textbf x)}\big)
-{\rm ln}\Gamma (k)\Big]H_{\epsilon}\Big(\phi(\textbf y)\Big)d\textbf yd\textbf x +\gamma_{2}\\
&\int_{\Omega}\!\!\int_{\Omega} K_{\tau}(\textbf x,\textbf y)
\!\Big[(\kappa -1){\rm ln}\big(I(\textbf y)\big)\!-\!\kappa {\rm ln}\big(\sigma_{i}(\textbf x)\big)\\
&-\big(\frac{\kappa I(\textbf y)}{\sigma_{i}(\textbf x)}\big)
-{\rm ln}\Gamma (k)\Big]\!\Big(\!1\!-\!H_{\epsilon}\big(\phi(\textbf y)\!\big)\!\Big)\!d\textbf yd\textbf x\\
=&\!\!\sum_{i=1}^{2}\!\!\gamma_{i}\!\!\int_{\Omega}\!\!\int_{\Omega}\!\!\!\! K_{\tau}(\textbf x,\textbf y)\!\Big[\!(\kappa \!-\!\!1){\rm ln}\big(I(\textbf y)\big)\!-\! \kappa {\rm ln}\big(\sigma_{i}(\textbf x)\big)\\
&-\big(\frac{\kappa I(\textbf y)}{\sigma_{i}(\textbf x)}\big)
-{\rm ln}\Gamma (k)\!\Big]
M_{i}^{\epsilon}\Big(\phi(\textbf y)\Big)d\textbf yd\textbf x
\end{split}
\end{equation}
Eqs. (\ref{minimization for oil features}) and (\ref{intermidiate term}) have been generated for the minimisation of the segmentation energy functional, and thus under the Gamma distribution, these two formulas are revised as:
\begin{equation}
\begin{split}
\int K_{\tau}\big(\textbf x,\textbf y\big)
\Big[M_{i}^{\epsilon}\big(\phi(\textbf y)\big)I(\textbf y)\Big]d\textbf y-&\int K_{\tau}\big(\textbf x,\textbf y\big)\sigma_{i}\big(\textbf x\big)\\
&M_{i}^{\epsilon}\Big(\phi(\textbf y)\Big)d\textbf y=0
\end{split}
\end{equation}
and 
\begin{equation}
\begin{split}
\widetilde{\varepsilon _{g_{i}}}\big(\textbf x\big)=\!\!\int\!\! K_{\tau}\big(\textbf y,\textbf x\big)&\Big[(\kappa-1){\rm ln}\big(I(\textbf x)\big)-\kappa{\rm ln}\big(\sigma_{i}(\textbf y)\big)\\
&-\big(\frac{\kappa I(\textbf x)}{\sigma_{i}(\textbf y)}\big)-{\rm ln}\Gamma(\kappa)\Big]d\textbf y
\end{split}
\end{equation}
So far, we have given the derivations of Eqs. (\ref{the energy function without constant}), (\ref{level set energy functional for fitness}), (\ref{minimization for oil features}) and (\ref{intermidiate term}) in terms of the Weibull and Gamma distributions. From these equations, it is observed that under different distributions, the energy functional embedded in our proposed segmentation algorithm can be adapted to account for the deployed distribution models for image segmentation.

\section{Experimental work}
In this section, we perform experimental validation by comparing our proposed method against several state-of-the-art segmentation methodologies, and the visual and quantitative experimental results are shown in the following parts.

Specifically, to conduct a comprehensive validation for our proposed method, we use a variety of SAR images with VH, VV and HH polarization obtained from the NOWPAP database\footnote{http://cearac.poi.dvo.ru/en/db/}, as our test dataset. Particularly, the oil spill images used in our experiments include C-band SAR images from the ERS-1 and ERS-2 satellites, and C-band ASAR images from the Envisat satellite. These images contain oil spills with various areas and irregular shapes, captured in different times by different sensors. We summarise the detailed information of SAR image sources and sensor properties in Tables \ref{SAR image information of NOWPAP} and \ref{information of SAR sensors}, where the symbol ``-" indicates unavailable information. In addition, for the oil spill SAR images used in the experimental part, the corresponding ground-truth source is also obtained from the NOWPAP dataset on the website. It was reported on the website that the ground-truth oil slick areas were generated by trained human experts with the usage of the commercial “Photoshop” package. The external information they used includes the visual spill properties such as shape and contrast between the feature and the surrounding sea, the presence and location of oil platforms and other stationary objects relative to the dark features, and the supplementary information such as wind speed, wave height, currents and so on. For example, the shape of oil spills depends on the wind speed, and the current history. Besides, the procedure they used normally includes  adaptive thresholding, and after the initial thresholding, a clustering step is performed. Specifically, adaptation is necessary because of the radar contrast between the dark patch and the surrounding sea. Moreover, the clustering step is performed to check the border between the dark patch and the surrounding sea. For the oil spill SAR images exploited in the experimental part, they have the same oil spill thickness: 0.2 $\mu$m. However, the oil spill areas and oil spill volumes are different. Specifically, for ERS-1 SAR, ERS-2 SAR and Envisat ASAR images, the oil spill areas are 394 km$^{2}$, 42.6x$10^{6}$m$^{2}$ and  53x$10^{6}$m$^{2}$ respectively, and the oil spill volumes are 78.8 m$^{3}$, 8.5 m$^{3}$ and $>$10 m$^{3}$ separately. The code of our proposed method is available at\footnote{https://github.com/FangChen160/pdmoilspillseg}.

\begin{table}[htbp]
	\renewcommand\arraystretch{2.0}
	\centering
	\tabcolsep 0.023in
	\caption{NOWPAP SAR IMAGE DESCRIPTIONS.}
	\begin{tabular}{c|c|c|c}
		\hline
		\hline
		Capture Time & Satellite & Type of Oil Spills & Image Cover Ground \\
		\hline
		19.06.1995 02:30:12 & ERS-1 &  - & 394 km$^{2}$  \\
		\hline
		08.11.1993 01:46:25 & ERS-1 & -& -\\
		\hline
		08.11.1993 01:46:10 & ERS-1 & - & -\\
		\hline
		19.06.1995 02:30:55 & ERS-1 & - & -\\
		\hline
		20.07.1997 02:14:41 & ERS-2 & Ship Oil Spill & -  \\
		\hline
		17.12.1997 01:57:42 & ERS-2 & Ship Oil Spill &- \\
		\hline
		19.06.1995 02:30:12 & ERS-2 & Ship Oil Spill &- \\
		\hline
		02.09.1996 02:00:55 & ERS-2 & - &17.8x106m$^{2}$ \\
		\hline
		16.08.2007 01:16:02  & Envisat &  Ship Oil Spill &- \\
		\hline
		01.09.2008 01:11:51 & Envisat & - &-  \\
		\hline
		14.04.2004 01:28:00 & Envisat & - & -\\
		\hline
		05.08.2008 01:59:45 & Envisat & - & -\\
		\hline
		\hline
	\end{tabular}
	\label{SAR image information of NOWPAP}	
\end{table}

\begin{table}[htbp]
	\renewcommand\arraystretch{2.3}
	\centering
	\tabcolsep 0.1in
	\caption{SAR SENSORS AND IMAGE DESCRIPTIONS.}
	\begin{tabular}{c|c|c|c}
		\hline
		\hline
		Satellite& Spatial Resolution & Band & Image Level  \\
		\hline
		ERS-1,2 SAR & 30m x 30m  & C-band & 2\\
		\hline
		Envisat ASAR & 150m x 150m  &C-band & 2 \\
		\hline
		\hline
	\end{tabular}
	\label{information of SAR sensors}	
\end{table}

\subsection{Segmentation on Multiple Oil Spill Images}
\label{machine learning methods for different oil spill iamges segmentation}
\begin{figure*}
	\begin{center}
    \vspace*{-7mm}
	\includegraphics[width=1.2\textwidth,height=0.9\textheight,center]{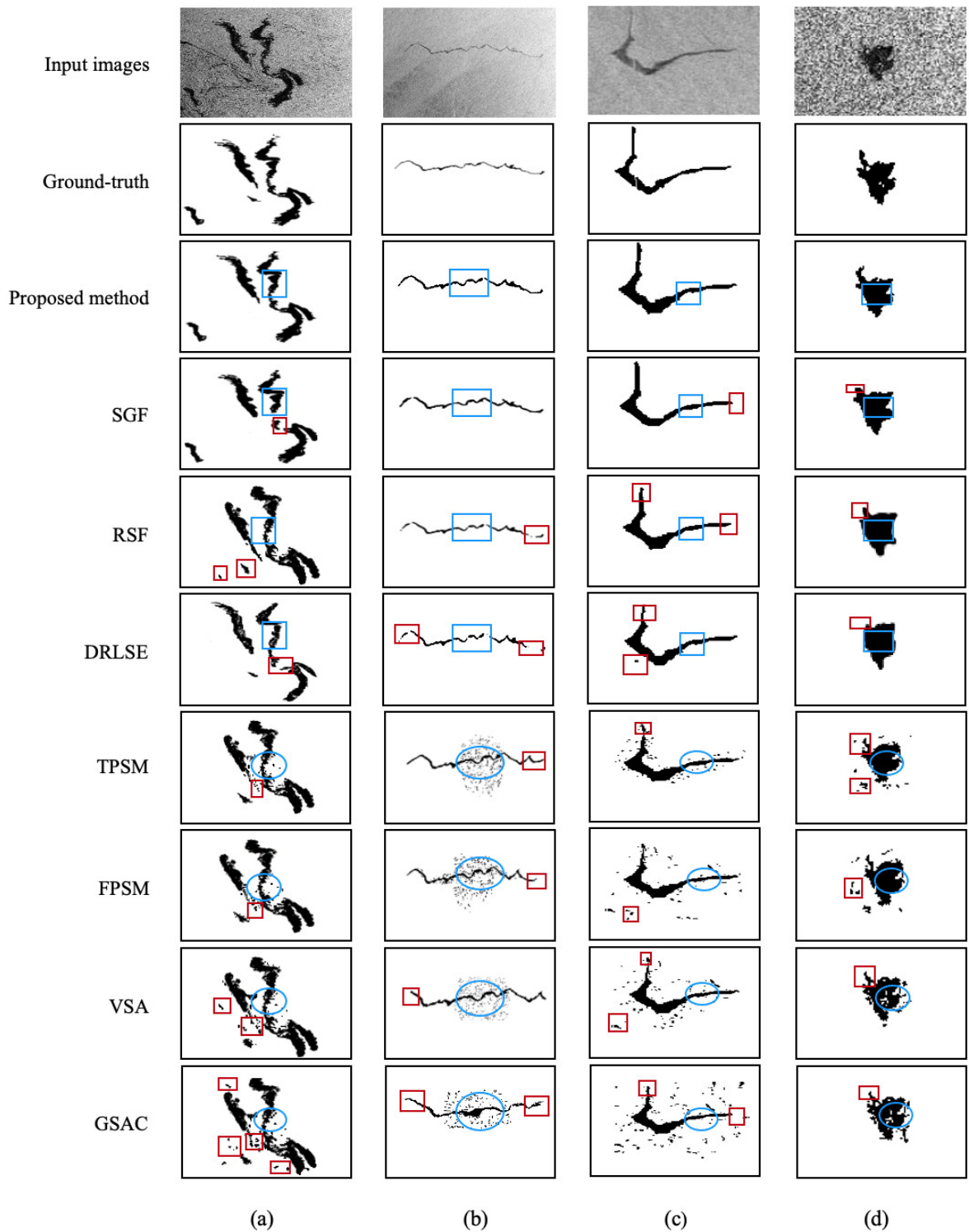}
	\end{center}
	\vspace*{-7mm}
	\caption{Oil spill segmentation on the ERS-1 oil spill SAR images. Specifically, (a) -(d) show the original oil spill SAR images and their related segmentation results by different segmentation techniques. Particularly, the blue rectangle boxes and the blue circles are exploited to indicate the initialisation for these implemented segmentation techniques, and the incorrect segmentation areas are indicated with red rectangle boxes.}
  \label{ERS-1 oil spill SAR iamges segmentation}
\end{figure*}

\begin{figure*}
	\begin{center}
    \vspace*{-7mm}
	\includegraphics[width=1.2\textwidth,height=0.9\textheight,center]{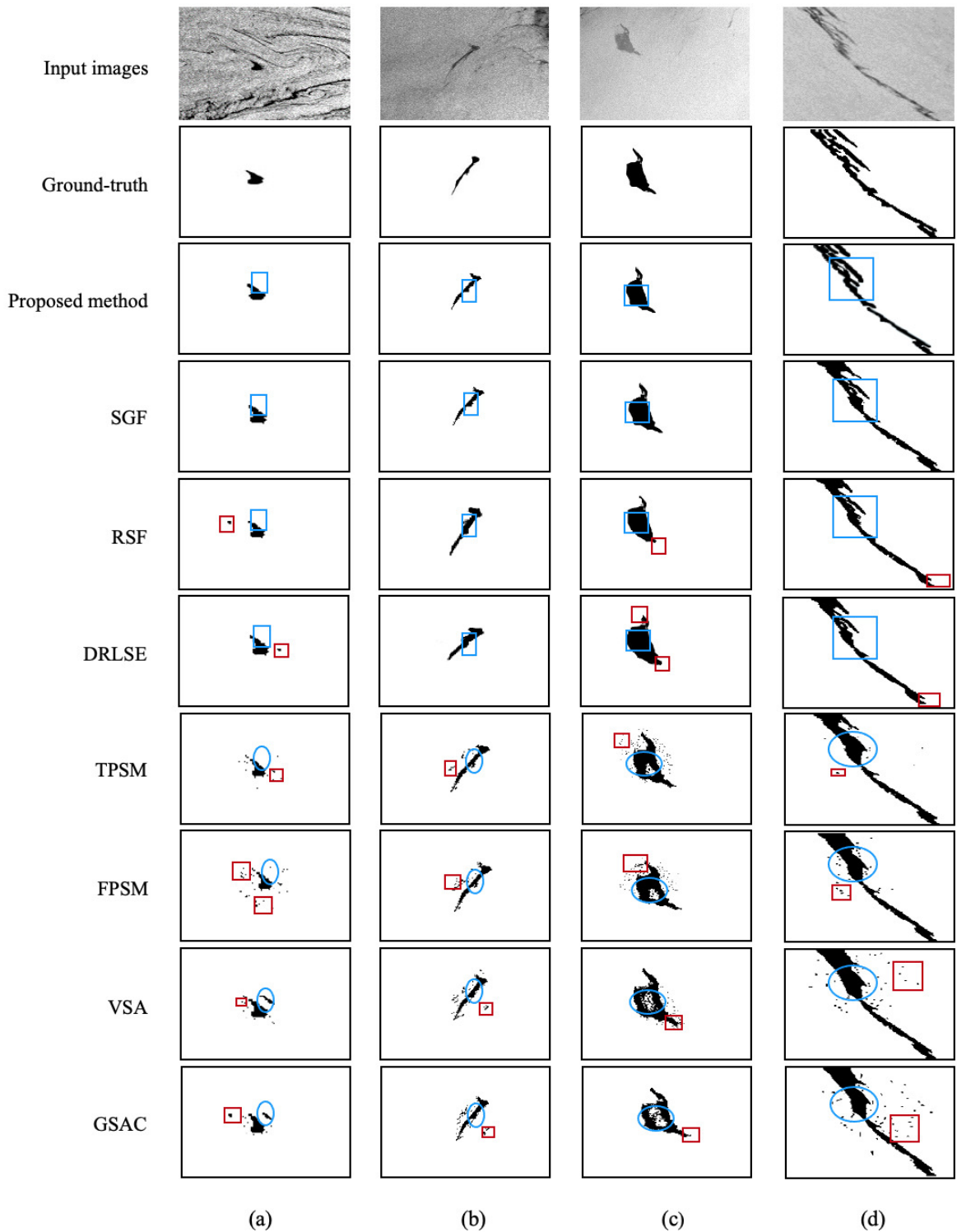}
	\end{center}
	\vspace*{-7mm}
	\caption{Oil spill segmentation results on the ERS-2 oil spill SAR images. In this figure, (a)-(d) show the original oil spill SAR images and their related segmentation results by different segmentation techniques. Particularly, the blue rectangle boxes and the blue circles are exploited to indicate the initialisation for these implemented segmentation techniques, and the incorrect segmentation regions are marked by red rectangle boxes.}
  \label{ERS-2 oil spill SAR image segmentation}
\end{figure*}

\begin{figure*}	
\begin{center}
 \vspace*{-7mm}
\includegraphics[width=1.2\textwidth,height=0.9\textheight,center]{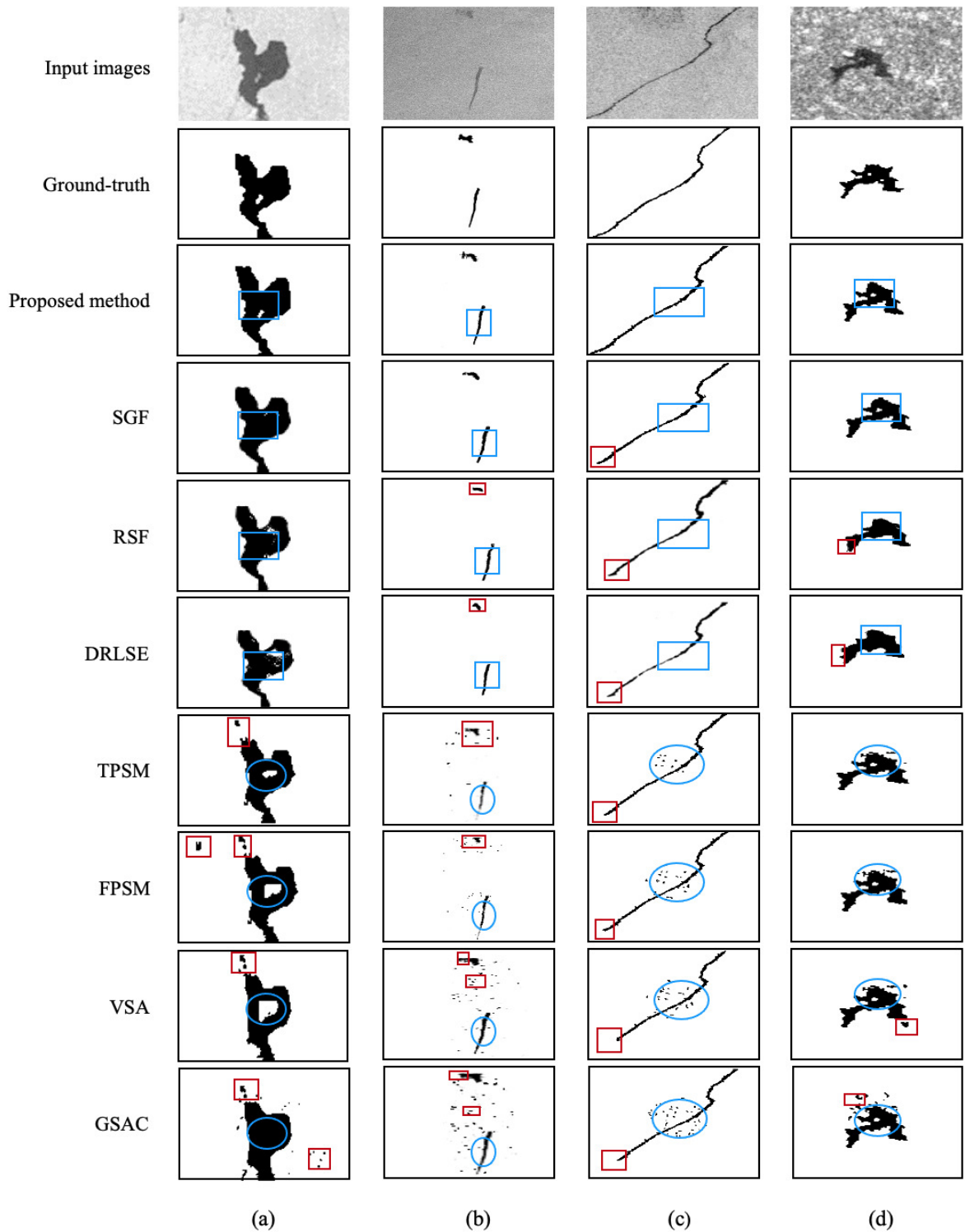}
\end{center}
\vspace*{-7mm}
\caption[]{Oil spill segmentation results on the Envisat ASAR images. Columns (a)-(d) are the original oil spill images and their related segmentations by our proposed method and the other comparison methods. Particularly, the blue rectangle boxes and the blue circles are exploited to indicate the initialisation for these implemented segmentation techniques, and the incorrect segmentation areas are illustrated with red rectangle boxes.}
\label{envisat oil spill SAR image segmentation}	
\end{figure*}

In order to conduct comprehensive validations for our proposed method, in this part, we address the segmentation evaluations with different types of oil spill SAR images by comparing our proposed method against several other state-of-the-art segmentation methodologies. These comparison methods include the self-guided filtering (SGF) embedded method for edge preserving segmentation \cite{chen2017level}, the region-scalable fitting (RSF) method which focuses on the inhomogeneous intensity segmentation problem \cite{li2008minimization}, the distance regularisation level set evolution (DRLSE) method \cite{li2010distance}, two-phase segmentation model (TPSM)\cite{zhang2015level}, four-phase segmentation model (FPSM) \cite{zhang2015level}, variational-based segmentation approach (VSA) \cite{zhang2014variational} and global statistical active contour (GSAC) method \cite{song2013globally}. The segmentation results are shown in Figs. \ref{ERS-1 oil spill SAR iamges segmentation}-\ref{envisat oil spill SAR image segmentation}. Specifically, Figs. \ref{ERS-1 oil spill SAR iamges segmentation}, \ref{ERS-2 oil spill SAR image segmentation} and \ref{envisat oil spill SAR image segmentation} illustrate the segmentation results on the ERS-1 oil spill SAR, the ERS-2 oil spill SAR, and the Envisat oil spill ASAR images, respectively. It is observed that our proposed method generates accurate oil spill segmentations.

The visual segmentation results demonstrate the superior performance of our proposed method. To conduct a more comprehensive evaluation for the proposed method, quantitative evaluations are also conducted, and we perform the accuracy calculation as follows:
\begin{equation*}\label{accuracy computation}
\text{Accuracy} = \frac{  \text{\# the number of correctly segmented pixels}}{ \text{\# the number of all pixels}}
\end{equation*}
which manifests a pixel-level evaluation for the segmentation task. The computed segmentation accuracy is shown in Tables \ref{ERS-1 oil spill SAR image segmentation accuracy}-\ref{Envisat oil spill SAR image segmentation accuracy}.

Specifically, Table \ref{ERS-1 oil spill SAR image segmentation accuracy} shows the segmentation accuracy of the implemented methodologies on ERS-1 oil spill SAR image segmentation, Table \ref{ERS-2 oil spill SAR image segmentation accuracy} shows the segmentation accuracy of the same methods used for ERS-2 oil spill SAR image segmentation, and Table \ref{Envisat oil spill SAR image segmentation accuracy} shows the segmentation accuracy with the exploitation of the implemented methods on Envisat oil spill ASAR image segmentation. From these Tables, it is observed that among these comparison methods, the category of image domain transformation techniques (i.e. GSAC, VSA, FPSM and TPSM) achieve less satisfactory oil spill segmentation. This is because these methods perform the segmentation on the transformed image domain and the transformation operation causes the original image information to be lost to some extent, and the reduction of the image information degenerates the segmentation. The active contour oriented techniques (i.e. DRLSE, RSF, SGF) perform the segmentation by introducing active contour models to characterise the region or edge information of the oil spill images without transformation. Therefore, the segmentation procedure is operated on the original images without information loss, and thus achieves higher segmentation accuracy. In contrast, our developed method achieves the highest segmentation accuracy. This performance is beneficial from the comprehensive consideration of SAR image formation which describes the oil spills in SAR images and oil spill segmentation simultaneously. These quantitative validation results are well in line with the visual validation results shown in Figs. \ref{ERS-1 oil spill SAR iamges segmentation}, \ref{ERS-2 oil spill SAR image segmentation} and \ref{envisat oil spill SAR image segmentation}.

The segmentation accuracy describes the segmentation performance in terms of the correctly segmented pixels against the overall pixels. In order to evaluate the segmentation performance further, we conduct the precision evaluation for the  proposed method and the other comparison methodologies. The precision evaluation is performed by the correct segmentation of the oil spills and the all segmentations in which the seawater that detected as oil spills is included. This gives a more specific demonstration for the oil spill segmentation, and the precision is calculated as follows:
\begin{equation*}
\text{Precision}\!\! = \!\!\frac{\!\!\text{\# the number of correctly segmented oil spill pixels}}{ \text{\# the number of segmented pixels}}
\end{equation*}
The calculation results with respect to different types of oil spill image segmentation are shown in Tables \ref{Precision of ERS-1 oil spill SAR image segmentation}-\ref{Precision of Envisat oil spill ASAR image segmentation}. Specifically, Tables \ref{Precision of ERS-1 oil spill SAR image segmentation}, \ref{Precision of ERS-2 oil spill SAR image segmentation} and \ref{Precision of Envisat oil spill ASAR image segmentation} show the segmentation precision of ERS-1 , ERS-2 and Envisat oil spill images respectively. From these Tables, it is clear that our proposed method also achieves the most accurate oil spill segmentation in terms of precision.

\begin{figure*}
  \begin{center}
  %
  \subfigure[]{
  \includegraphics[width=0.31\textwidth,height=0.18\textheight]{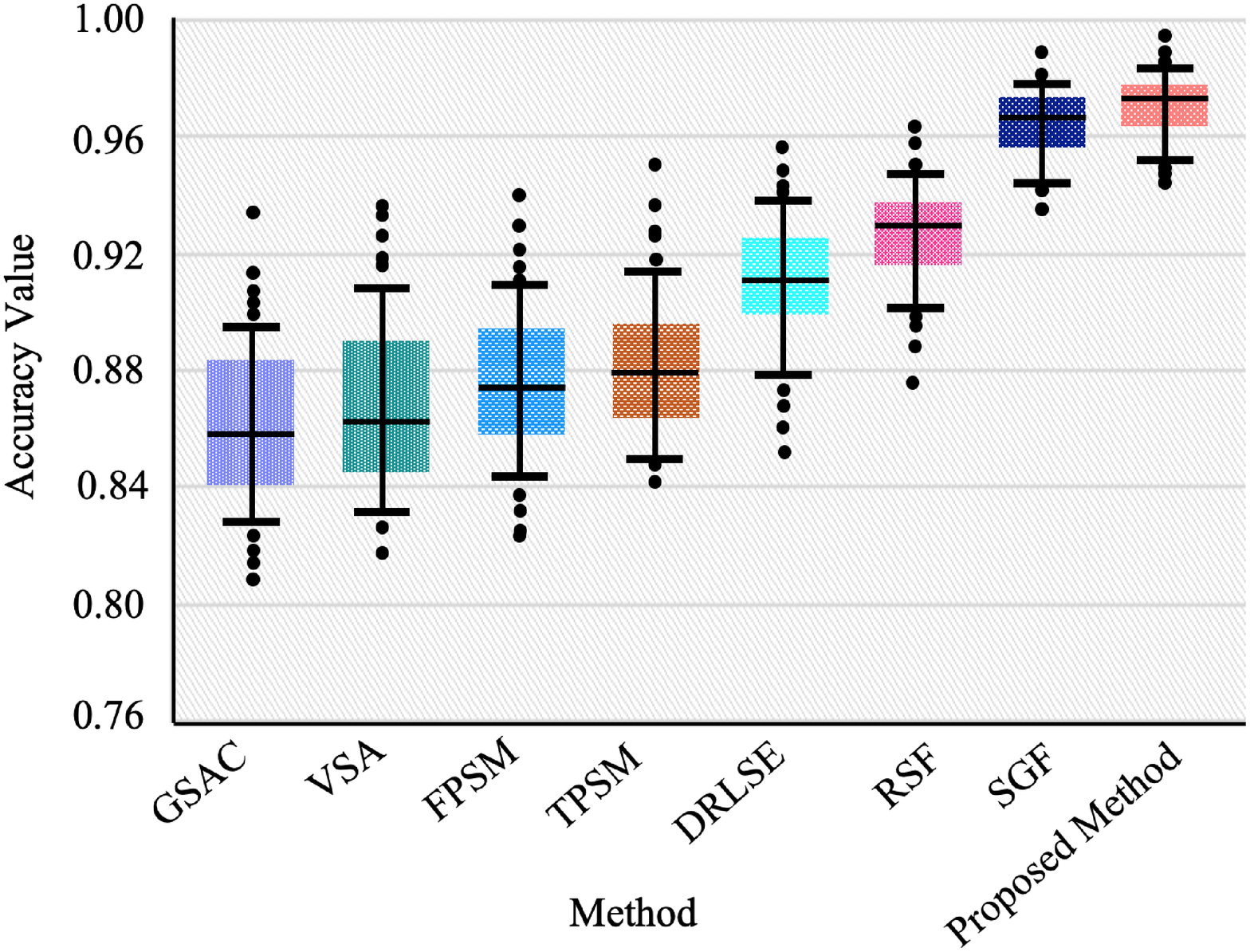}}
  \subfigure[]{
  \includegraphics[width=0.31\textwidth,height=0.18\textheight]{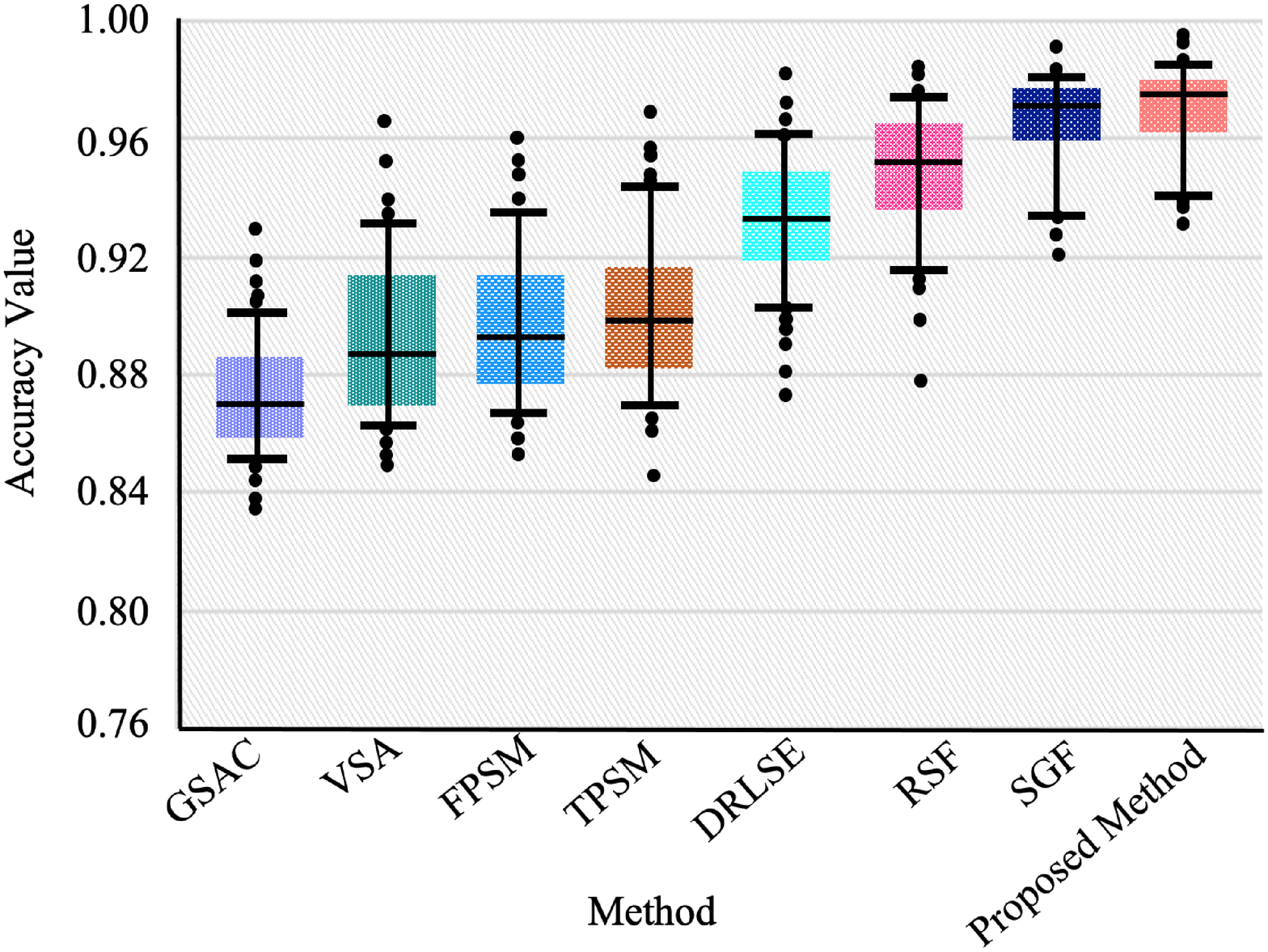}}
  \subfigure[]{
  \includegraphics[width=0.31\textwidth,height=0.18\textheight]{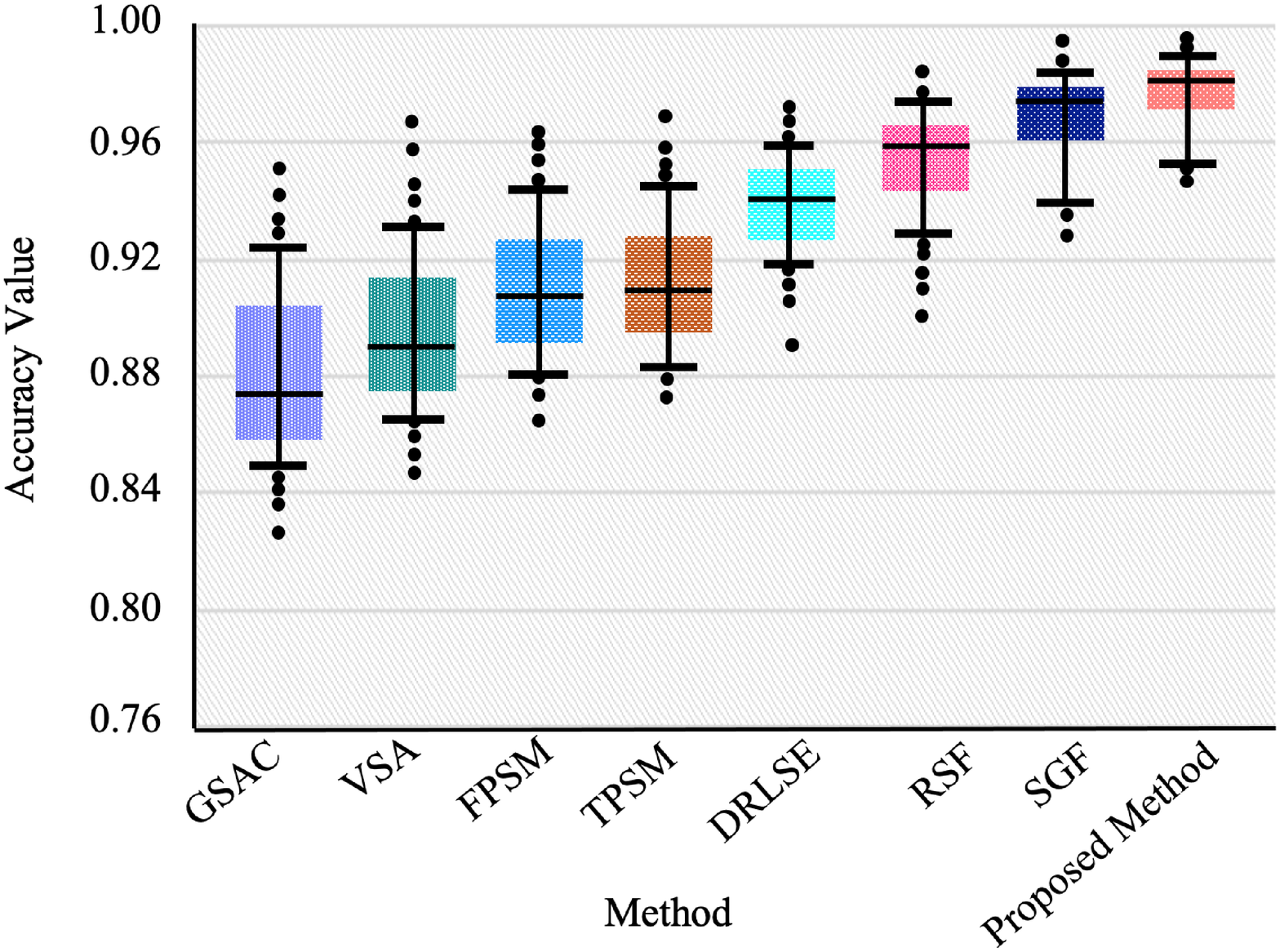}}
  \end{center}
  \vspace*{-4mm}
  \caption{Segmentation accuracy distributions with respect to different types of oil spill SAR image segmentation. Specifically, (a), (b) and (c) illustrate the distributions of the segmentation accuracy that corresponds to each method for ERS-1, ERS-2 and Envisat oil spill image segmentation, respectively. The rectangles are exploited to depict the accuracy distributed between the minimum and maximum values (these two values are shown with bars on the bottom and top of the rectangles), and the black dots are the segmentation outliers that fall out the interval of the two bars.}
  \label{accuracy with standard deviation}
\end{figure*}

In addition, to give a more comprehensive validation for our proposed method, we conduct the segmentation over 200 SAR and ASAR images, and the statistical results for reflecting the overall performance are shown in Fig. \ref{accuracy with standard deviation}. Here, we employ the box plots to measure the effectiveness of these implemented methods in terms of their segmentation accuracy and stability. Specifically, Fig. \ref{accuracy with standard deviation} (a), (b) and (c) show the segmentation accuracy distribution ranges of ERS-1, ERS-2 and Envisat oil spill images respectively. In each sub-figure, the colored boxes depict the scope of the clustered accuracy values, where the bars on the bottom and up of the boxes are the levels of the minimum and maximum values, and the black dots are the segmentation outliers that fall out the segmentation convergence intervals. Overall, our proposed method achieves the highest segmentation accuracy with the smallest variation.

\begin{figure*}
  \begin{center}
  %
  \subfigure[]{
  \includegraphics[width=0.466\textwidth,height=0.26\textheight]{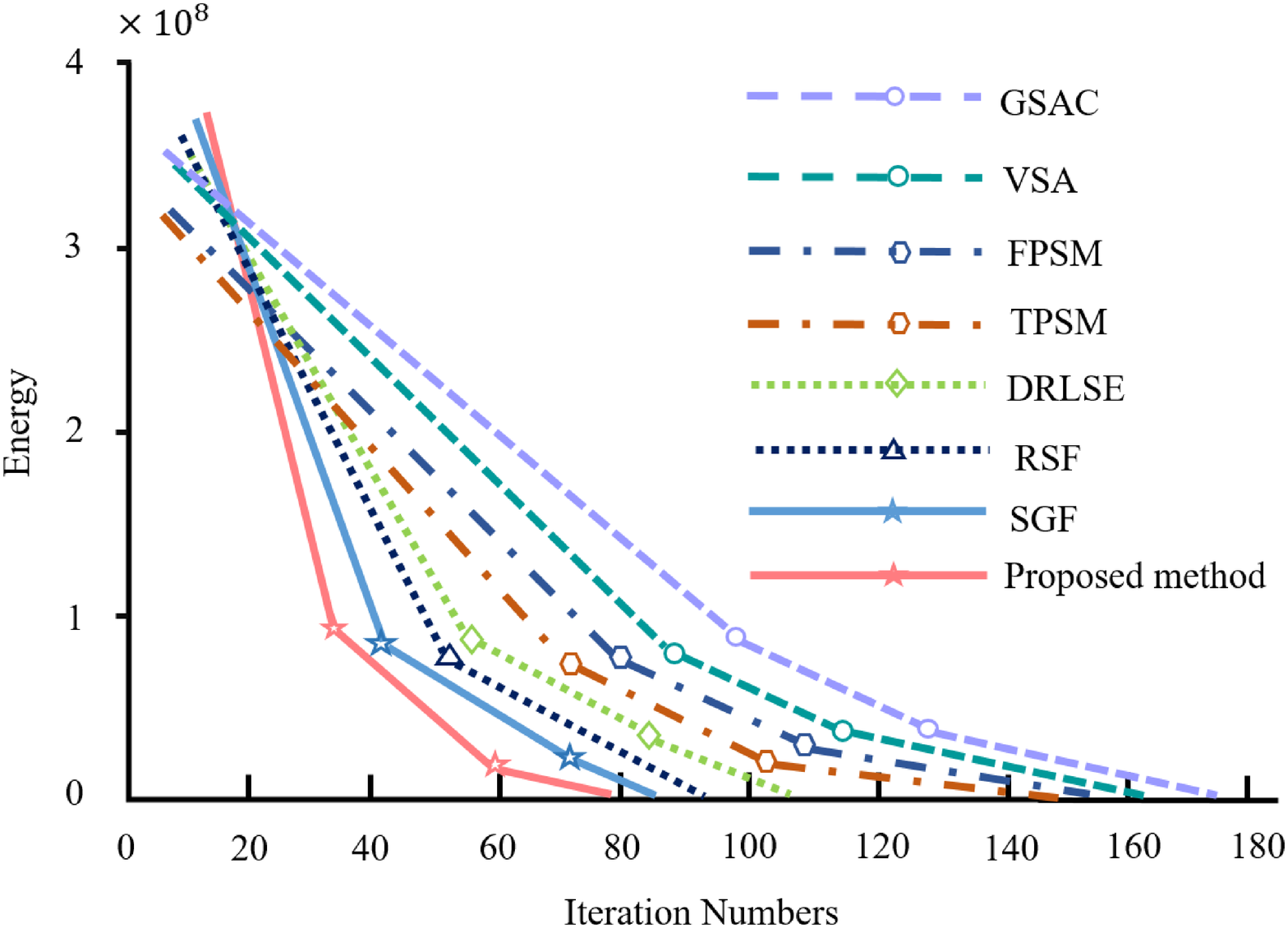}}
  \subfigure[]{
  \includegraphics[width=0.46\textwidth,height=0.26\textheight]{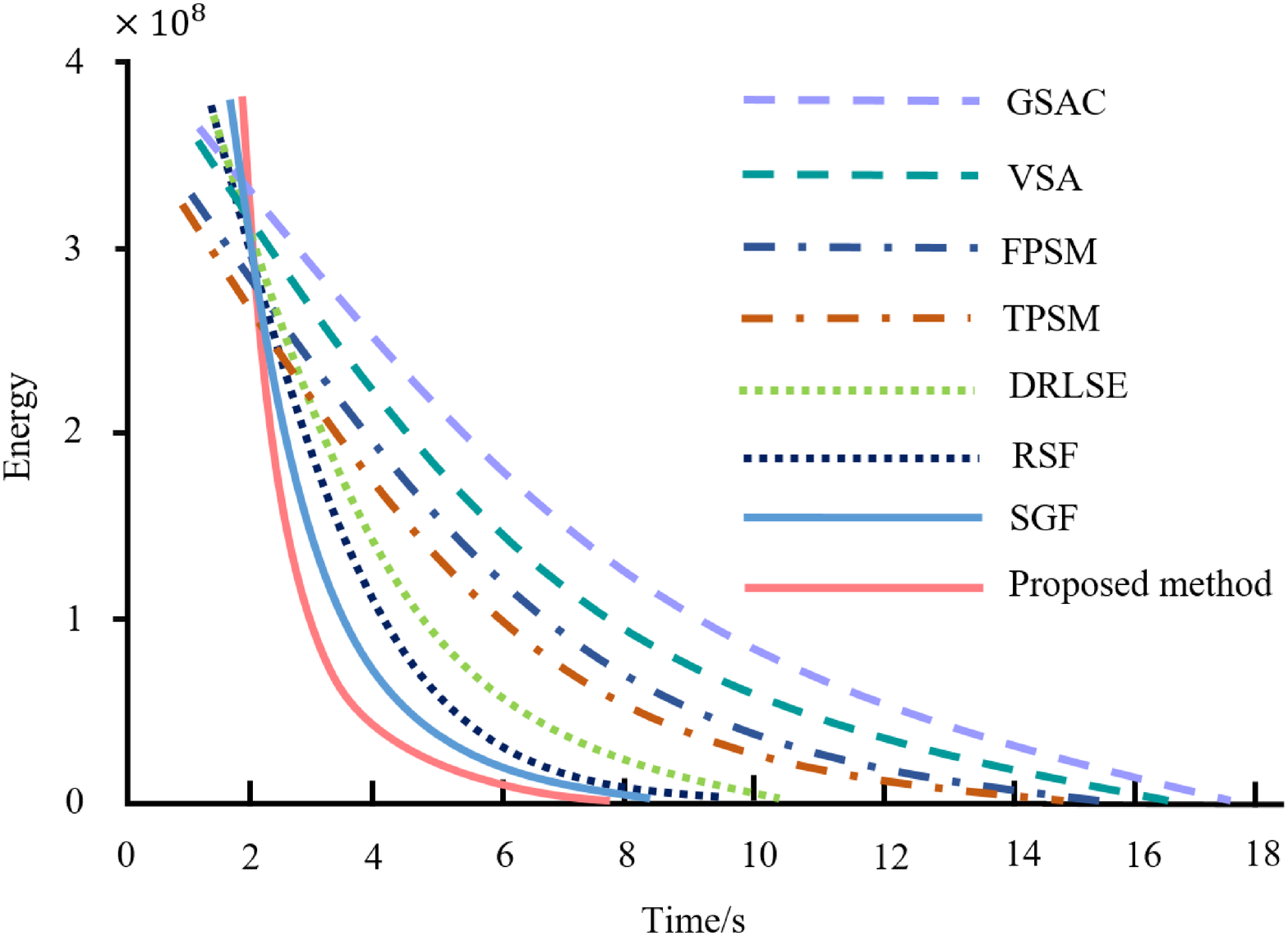}}
  \end{center}
  \vspace*{-4mm}
  \caption{Segmentation performance in terms of iteration numbers and runtime. Specifically, (a) and (b) demonstrate the segmentation evolution rate with respect to iteration numbers and runtime, respectively.}
  \label{convergnece rates with respect to iteration numbers and runtime}
\end{figure*}

To make a further evaluation for these segmentation methods, we validate the segmentation performance in terms of iteration numbers and runtime, which play a key role on evaluating their systematic performance, and the results are shown in Fig. \ref{convergnece rates with respect to iteration numbers and runtime}. Specifically, Fig. \ref{convergnece rates with respect to iteration numbers and runtime} (a) and (b) illustrate the energy minimisation evolution rate of these methods with respect to iteration numbers and runtime respectively under Matlab 2018a implementation based on Intel(R) Core(TM) i7-8700CPU@3.20GHz 3.19 GHz. These results demonstrate that our proposed method performs the most efficient oil spill segmentation with better convergence.  Particularly, from these results, we obtain that the convergence rate of the proposed method is almost over a half faster than the average rate of the other comparison methods for both iteration numbers and runtime. This is beneficial from the incorporation of the image distribution model in which the backscattering coefficient helps to characterise the oil spills in the SAR images. Therefore, the distribution model encourages the segmentation procedure to operate targeting at the oil spill areas during the segmentation evolution and thus achieves better convergence.

\begin{table*}
	\renewcommand\arraystretch{2.6}
	\begin{center}
	\tabcolsep 0.1023in
	\caption{OIL SPILL SEGMENTATION ACCURACY FOR THE ERS-1 SAR IMAGES.}
	\begin{tabular}{c|cccc|ccc|c}
		\hline
		\hline
		\multirow{2}{*}{\textbf{Image Number}}&\multicolumn{4}{c|}{\textbf{Image Domain Transformation Technique}} &\multicolumn{3}{c|}{\textbf{Active Contour Oriented Method}}&\textbf{Novel Method}\\
		\cline{2-9}
		&\textbf{GSAC} & \textbf{VSA} & \textbf{FPSM} & \textbf{TPSM} &\textbf{DRLSE} & \textbf{RSF} & \textbf{SGF} & \textbf{The Proposed Method}\\
		\hline
		\hline
		(a)& $0.8891$ & $0.8926$ & $0.9019$ & $0.9075$ & $0.9306$ & $0.9423$ & $0.9519$ & $0.9685$\\
		\hline
	    (b)& $0.8496$ & $0.8429$ & $0.8579$ & $0.8602$ & $0.9521$ & $0.9602$ & $0.9791$ & $0.9816$\\
		\hline
		(c) & $0.8529$ & $0.8668$ & $0.8793$ & $0.8975$ & $0.9295$ & $0.9286$ & $0.9865$ & $0.9906$\\
		\hline
		(d) & $0.8608$ & $0.8795$ & $0.8897$ & $0.9036$ & $0.9021$ & $0.9429$ & $0.9576$ & $0.9880$\\
		\hline
		\hline
	\end{tabular}
	\label{ERS-1 oil spill SAR image segmentation accuracy}
	\end{center}
\end{table*}

\begin{table*}
	\renewcommand\arraystretch{2.6}
	\begin{center}
	\tabcolsep 0.1023in
	\caption{OIL SPILL SEGMENTATION ACCURACY FOR THE ERS-2 SAR IAMGES.}
	\begin{tabular}{c|cccc|ccc|c}
		\hline
		\hline
		\multirow{2}{*}{\textbf{Image Number}}&\multicolumn{4}{c|}{\textbf{Image Domain Transformation Technique}} &\multicolumn{3}{c|}{\textbf{Active Contour Oriented Method}}&\textbf{Novel method}\\
		\cline{2-9}
		&\textbf{GSAC} & \textbf{VSA} & \textbf{FPSM} & \textbf{TPSM} &\textbf{DRLSE} & \textbf{RSF} & \textbf{SGF} & \textbf{The Proposed Method}\\
		\hline
		\hline
		(a)& $0.8986$ & $0.9143$ & $0.9219$ & $0.9246$ & $0.9721$ & $0.9716$ & $0.9885$ & $0.9983$\\
		\hline
	    (b)& $0.9016$ & $0.9135$ & $0.9296$ & $0.9301$ & $0.9673$ & $0.9685$ & $0.9875$ & $0.9901$\\
		\hline
		(c) & $0.8352$ & $0.8765$ & $0.8892$ & $0.8926$ & $0.9375$ & $0.9386$ & $0.9836$ & $0.9879$\\
		\hline
		(d) & $0.8923$ & $0.8967$ & $0.9063$ & $0.9089$ & $0.9586$ & $0.9604$ & $0.9658$ & $0.9679$\\
		\hline
		\hline
	\end{tabular}
    \label{ERS-2 oil spill SAR image segmentation accuracy}
    \end{center}
\end{table*}

\begin{table*}
	\renewcommand\arraystretch{2.6}
	\begin{center}
	\tabcolsep 0.1023in
	\caption{OIL SPILL SEGMENTATION ACCURACY FOR THE ENVISAT ASAR IMAGES.}
	\begin{tabular}{c|cccc|ccc|c}
		\hline
		\hline
		\multirow{2}{*}{\textbf{Image Number}}&\multicolumn{4}{c|}{\textbf{Image Domain Transformation Technique}} &\multicolumn{3}{c|}{\textbf{Active Contour Oriented Method}}&\textbf{Novel Method}\\
		\cline{2-9}
		&\textbf{GSAC} & \textbf{VSA} & \textbf{FPSM} & \textbf{TPSM} &\textbf{DRLSE} & \textbf{RSF} & \textbf{SGF} & \textbf{The Proposed Method}\\
		\hline
		\hline
		(a)& $0.8365$ & $0.8663$ & $0.8795$ & $0.8749$ & $0.9319$ & $0.9476$ & $0.9623$ & $0.9962$\\
		\hline
	    (b)& $0.8602$ & $0.8955$ & $0.9079$ & $0.9052$ & $0.9803$ & $0.9819$ & $0.9826$ & $0.9982$\\
		\hline
		(c) & $0.8993$ & $0.9015$ & $0.9192$ & $0.9745$ & $0.9845$ & $0.9865$ & $0.9868$ & $0.9979$\\
		\hline
		(d) & $0.8975$ & $0.9029$ & $0.9673$ & $0.9682$ & $0.9739$ & $0.9785$ & $0.9906$ & $0.9921$\\
		\hline
		\hline
	\end{tabular}
    \label{Envisat oil spill SAR image segmentation accuracy}
    \end{center}
\end{table*}

\begin{table}[h]
  \renewcommand\arraystretch{2.3}
  \centering
  \tabcolsep 0.09in
  \caption{PRECISION OF THE ERS-1 OIL SPILL SAR IMAGE SEGMENTATION.}
  \begin{tabular}{cc|cccc}
    \hline
    \hline
    \multicolumn{2}{c|}{\diagbox[width=3.4cm,dir=NW]{Method}{Precision}{Image}} & (a) & (b) & (c) & (d)  \\
    \hline
    The Proposed Method  && 0.9774 & 0.9869 & 0.9958 & 0.9961 \\
    \hline
    SGF&& 0.9583 & 0.9806 & 0.9879 & 0.9602 \\
    \hline
    RSF&& 0.9509 & 0.9647 & 0.9291 & 0.9523 \\
    \hline
    DRLSE&& 0.9347 & 0.9535 & 0.9306 & 0.9052 \\
    \hline
    TPSM&& 0.9068 & 0.8549 & 0.8968 & 0.9017 \\
    \hline
    FPSM&& 0.9059 & 0.8536 & 0.8751 & 0.8723 \\
    \hline
    VSA&& 0.8819 & 0.8423 & 0.8652 & 0.8687 \\
    \hline
    GSAC&& 0.8806 & 0.8417 & 0.8504 & 0.8628 \\
    \hline
    \hline
  \end{tabular}
  \label{Precision of ERS-1 oil spill SAR image segmentation}
\end{table}

\begin{table}[h]
 \renewcommand\arraystretch{2.3}
  \centering
  \tabcolsep 0.09in
  \caption{PRECISION OF THE ERS-2 OIL SPILL SAR IMAGE SEGMENTATION.}
  \begin{tabular}{cc|cccc}
    \hline
    \hline
    \multicolumn{2}{c|}{\diagbox[width=3.7cm,dir=NW]{Method}{Precision}{Image}} & (a) & (b) & (c) & (d)  \\
    \hline
    The Proposed Method && 0.9989 & 0.9931 & 0.9896 & 0.9708 \\
    \hline
    SGF&& 0.9902 & 0.9917 & 0.9869 & 0.9695 \\
    \hline
    RSF&& 0.9723 & 0.9701 & 0.9372 & 0.9674 \\
    \hline
    DRLSE&& 0.9727 & 0.9689 & 0.9369 & 0.9668 \\
    \hline
    TPSM&& 0.9207 & 0.9295 & 0.8916 & 0.9052\\
    \hline
    FPSM&& 0.9206 & 0.9274 & 0.8803 & 0.9025 \\
    \hline
    VSA&& 0.9109 & 0.9124 & 0.8757 & 0.8958 \\
    \hline
    GSAC&& 0.8973 & 0.9001 & 0.8349 & 0.8927 \\
    \hline
    \hline
  \end{tabular}
  \label{Precision of ERS-2 oil spill SAR image segmentation}
\end{table}

\begin{table}[h]
  \renewcommand\arraystretch{2.3}
  \centering
  \tabcolsep 0.09in
  \caption{PRECISION OF THE ENVISAT OIL SPILL ASAR IMAGE SEGMENTATION.}
  \begin{tabular}{cc|cccc}
    \hline
    \hline
    \multicolumn{2}{c|}{\diagbox[width=3.7cm,dir=NW]{Method}{Precision}{Image}} & (a) & (b) & (c) & (d)  \\
    \hline
    The Proposed Method && 0.9985 & 0.9987 & 0.9986 & 0.9923 \\
    \hline
    SGF&& 0.9647 & 0.9833 & 0.9898 & 0.9925 \\
    \hline
    RSF&& 0.9498 & 0.9839 & 0.9886 & 0.9801 \\
    \hline
    DRLSE&& 0.9346 & 0.9852 & 0.9879 & 0.9757 \\
    \hline
    TPSM&& 0.8753 & 0.9047 & 0.9743 & 0.9703 \\
    \hline
    FPSM&& 0.8805 & 0.9068 & 0.9185 & 0.9689 \\
    \hline
    VSA&& 0.8674 & 0.8945 & 0.9008 & 0.9031 \\
    \hline
    GSAC&& 0.8327 & 0.8549 & 0.8968 & 0.8966 \\
    \hline
    \hline
  \end{tabular}
  \label{Precision of Envisat oil spill ASAR image segmentation}
\end{table}

Additionally, we here describe the segmentation performance of the proposed method in terms of other interferences. For oil spill SAR images, low wind speeds, internal waves and rain cells produce low backscattering areas that bring challenges for the oil spill segmentations. Particularly, for the proposed segmentation method, the segmentation model is developed with the exploitation of the image probability distribution in which the coefficient regarding oil spills is considered. In addition, in the segmentation implementation, the procedure exploits a rectangle box to capture an initialization region for starting the segmentation, and the rectangle box gives prior information for the segmentation. Thus, the segmentation method is able to perform effective segmentation.

The experimental evaluations in this subsection have demonstrated the effectiveness of the proposed method. Specifically, in the experimental implementation for oil spill SAR image segmentation, it is observed that a lot of parameters are exploited in the segmentation procedure. For the parameters exploited in the proposed method, we examine to analyze their impact on the segmentation performance. This provides an efficient way for conducting oil spill image segmentation in terms of setting the segmentation parameters. Specifically, in the proposed method, the parameters include $K_{s}$, $\tau$, $\gamma _{1}$, $\gamma _{2}$, $\epsilon$, $\mu$ and $\nu$. Specifically, the parameter $K_{s}$ is the radar system constant, and it is related to the radar system configuration. Thus, in the segmentation operation, we define it as a stable constant. However, for the other parameters, they are introduced in the construction of the segmentation energy functional, and thus they take an impactive role for the segmentation performance. According to the experimental implementation, we find that the segmentation performance is not sensitive to the parameters $\tau$, $\mu$, and $\epsilon$, and thus for different oil spill image segmentation, these three parameters keep stable. In contrast, the parameters $\gamma_{1}$, $\gamma_{2}$ and $\nu$ are various for different oil spill SAR image segmentation. Particularly, the parameter $\nu$ ranges from $0.00007$ to $0.0004$ in the segmentation process, and $\gamma_{2}$ is greater than $\gamma_{1}$ and they are in the range from $2.3$ to $2.306$. To illustrate the impact of these three parameters on the segmentation performance, we describe the segmentation performance of the proposed method in terms of different parameter values in Table \ref{Impact of parameters on the segmentation performance}. In this table, for each parameter set, the mean accuracy is statistically calculated with the segmentation of over 50 oil spill images. From the mean accuracy values, we can have that for the parameters chosen from the given range, though the statistical mean accuracy of different oil spill image segmentation changes slightly, the proposed method performs effective oil spill segmentation. This demonstrates that in the operation of oil spill image segmentation, $\gamma_{1}$, $\gamma_{2}$ and $\nu$ should be set in the given range. In this regard, the segmentation performance is correlated to the technique parameter settings. Thus, in the experimental process, we setup the competing techniques for testing with appropriate parameters. Specifically, in the experimental part, the competing techniques are mainly categorised as the image domain transformation strategies and the active contour-oriented ones. Particularly, for image domain transformation strategies, the setup for them to implement the oil spill SAR image segmentation mainly includes the parameters of the kernel function, the initialization radius, the iteration numbers, and the coefficient of the total regularization term. For the active contour-oriented methods, the setup for them to implement the segmentation mainly includes the parameters of the fitting energy term, the coefficient of the zero level contour regularization term, the iteration numbers and the initialization size. Thus, for the setup of the competing techniques, we use different types of oil spill SAR images to train the segmentation techniques separately, and the appropriate parameters for the setup of the competing techniques are obtained to implement the optimal segmentations.

\begin{table*}[h]
\centering
\renewcommand\arraystretch{2.36}
	\tabcolsep 0.10in
	\caption{SEGMENTATION PERFORMANCE WITH RESPECT TO DIFFERENT PARAMETER VALUES.}
	\label{Impact of parameters on the segmentation performance}
	\begin{tabular}{c|c|c|c|c}
		\hline
		\hline
		\multirow{2}{*}{\diagbox[height=1.4cm,width=4.7cm,dir=NW]{Image}{Mean Accuracy}{Parameters}}&\textbf{$\nu=0.00007$} &\textbf{$\nu=0.00009$}&\textbf{$\nu=0.0002$}&\textbf{$\nu=0.0004$}\\
		\cline{2-5}
		&\textbf{$\gamma_{1}=2.3$, $\gamma_{2}=2.303$} & \textbf{$\gamma_{1}=2.3$,$\gamma_{2}=2.304$} & \textbf{$\gamma_{1}=2.3$,$\gamma_{2}=2.305$} & \textbf{$\gamma_{1}=2.3$,$\gamma_{2}=2.306$} \\
		\hline
		ERS-1& $0.9760$ & $0.9752$ & $0.9755$& $0.9759$\\
		\hline
	    ERS-2& $0.9876$ & $0.9865$ & $0.9794$& $0.9786$ \\
		\hline
		Envisat & $0.9917$ & $0.9866$ & $0.9858$& $0.9875$ \\
		\hline
		\hline
	\end{tabular}
\end{table*}

\subsection{Initialisation Dependence Evaluation for Oil Spill Segmentation}
In the implementation of oil spill SAR image segmentation, we observe that for the segmentation techniques such as local image fitting energy (LIFE) \cite{zhang2010active}, TPSM, RSF and DRLSE, different initialisation for starting the segmentation usually generates different outcomes. Thus, in this subsection, we conduct further evaluation for the segmentation performance of the proposed method by comparing its segmentation results with those of the initialisation dependent methods.

\begin{figure}[h]
  \flushright
  \subfigure[]{
  \includegraphics[width=0.57\textwidth,height=0.288\textheight]{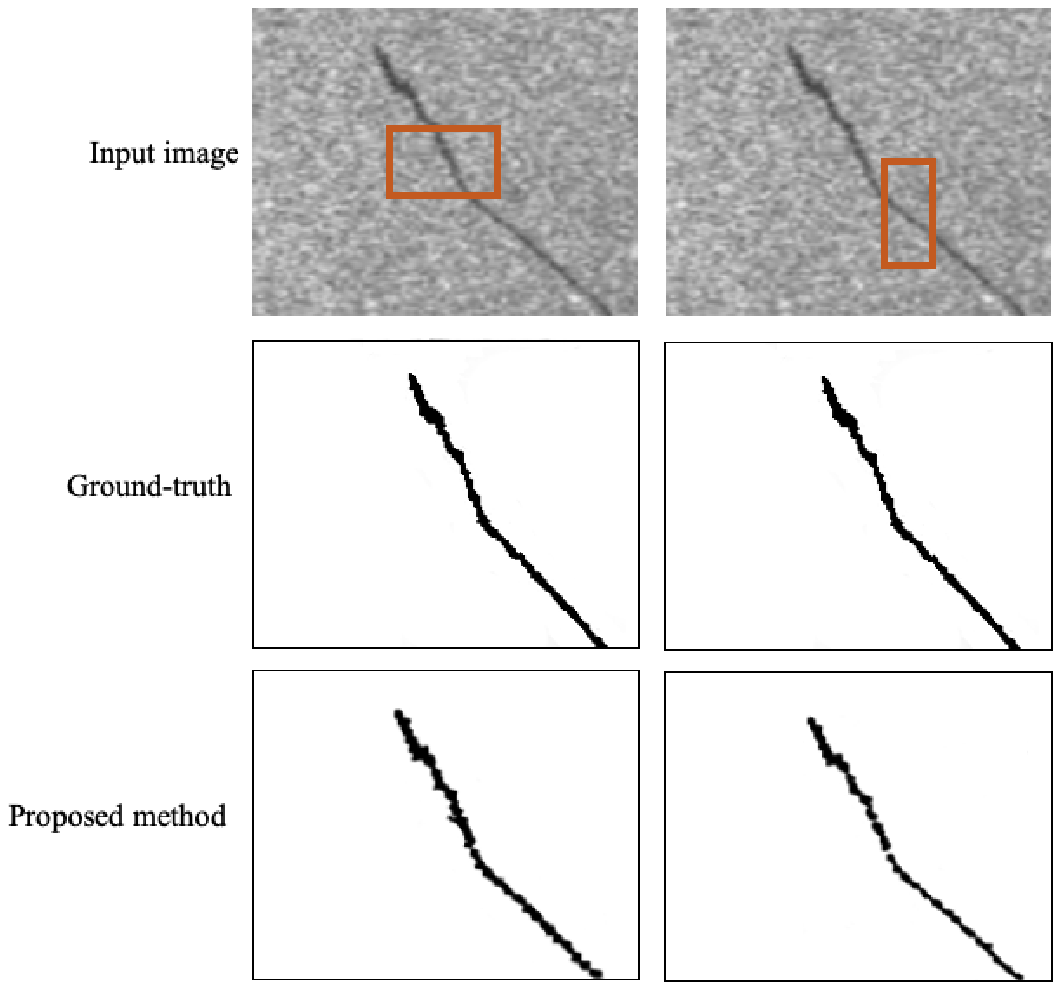}}
  \subfigure[]{
  \includegraphics[width=0.57\textwidth,height=0.288\textheight]{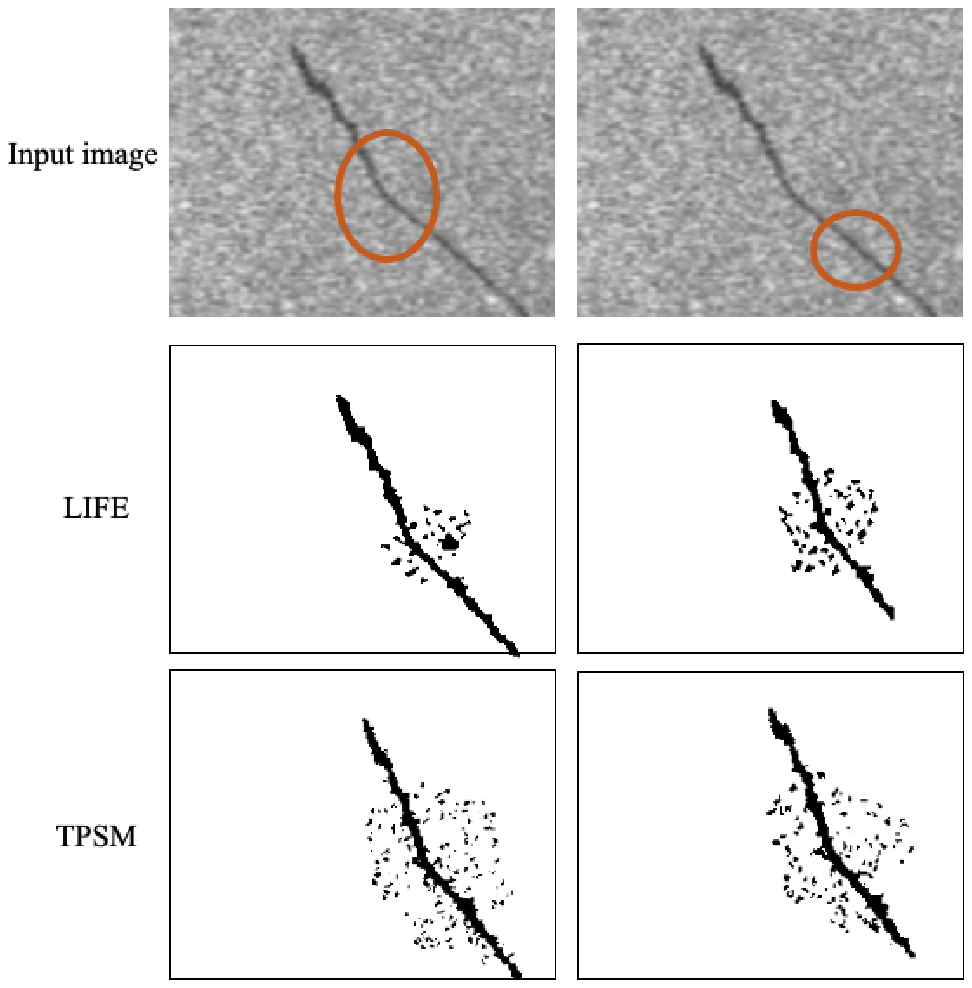}}
  \vspace*{-2mm}
  \caption{Comparison between the circular initialisation based strategies and the proposed method. Particularly, different line boxes at different locations are exploited to initialise for the proposed method, while different sized circles at different locations indicate the initialisation for LIFE and TPSM methods.}
  \label{circular radius initialization comparison}
\end{figure}

For these comparison methods (i.e. LIFE, TPSM, RSF and DRLSE), TPSM and LIFE techniques devise a circle to launch the segmentation, and the segmentation performance is affected by the initialisation location and the circle radius. In the meantime, RSF and DRLSE use a rectangle box to initialise the segmentation, which seems handy in practice. We commence the comparison by exploiting the proposed method and TPSM and LIFE techniques to segment the same oil spill SAR images with different sized circles to capture the initialisation regions, and the segmentation results are shown in Fig. \ref{circular radius initialization comparison}. In particular, Fig. \ref{circular radius initialization comparison} (a) and (b) show the segmentation results of the proposed method and the other two comparison methods respectively. It is clear that our proposed method achieves more accurate oil spill segmentation with different initialisations. This demonstrates that our proposed method performs more robust oil spill segmentation.

\begin{figure*}	
\begin{center}
\includegraphics[width=1.1\textwidth,height=0.7\textheight,center]{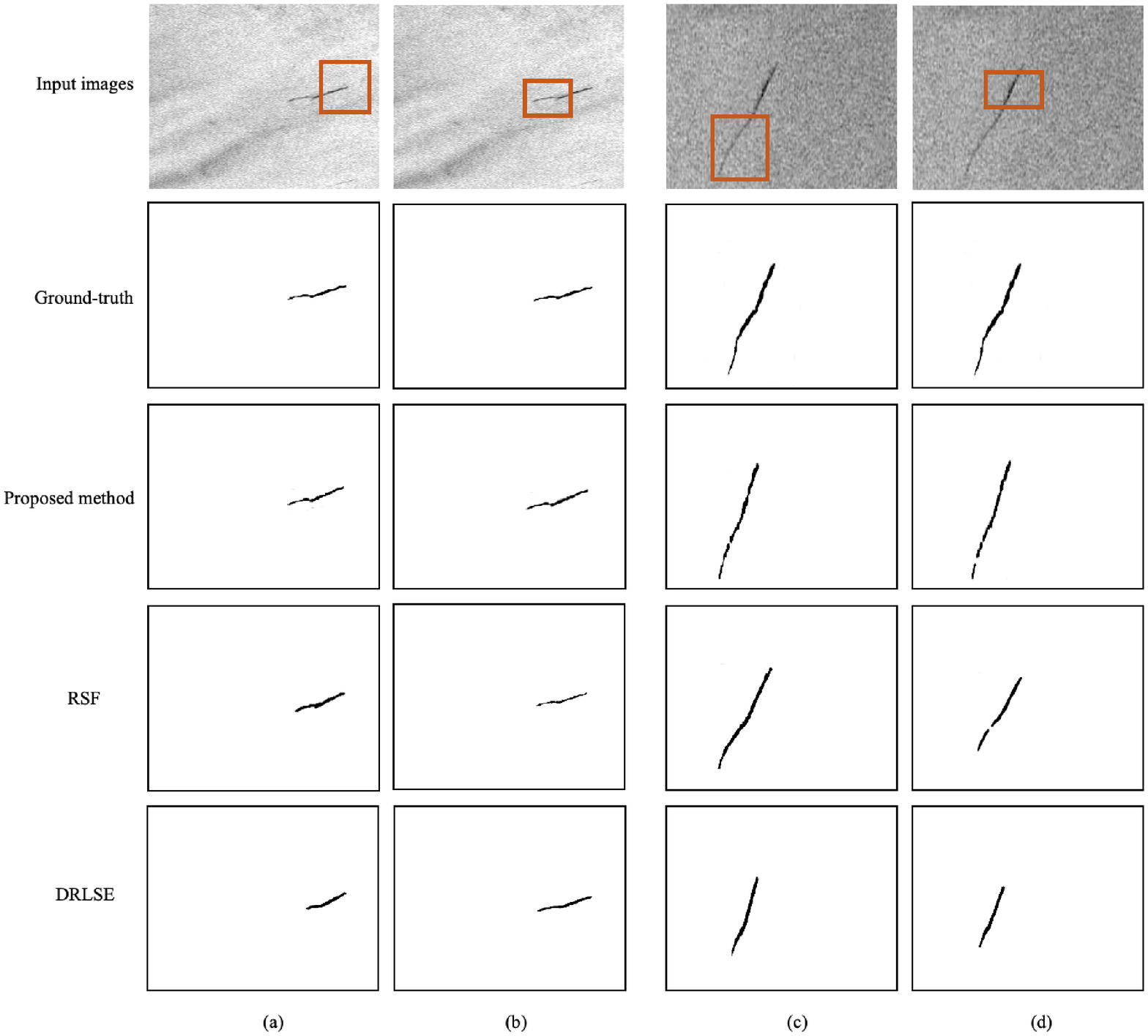}
\end{center}
\vspace*{-14mm}
\caption[]{Segmentation performance comparison between the proposed method and the rectangle box initialisation methods. From top to bottom of columns (a)-(d) are the original oil spill SAR images, the ground-truth, and the corresponding segmentation results of different segmentation methods. The rectangles in oil spill SAR images indicate different initialisations to start the oil spill segmentation.}
\label{line box initialization comparison}	
\end{figure*}

We then perform comparisons with the RSF and DRLSE segmentation methods. To undertake this comparison, different initialisations are used to capture the starting regions, and the segmentation results are shown in Fig. \ref{line box initialization comparison}. Carefully examining these segmentation results, we recognise that our proposed method achieves more stable oil spill segmentation with different initialisations, and its segmentation performance outperforms the RSF and DRLSE methods. This further manifests that our proposed method is capable of achieving robust oil spill segmentation in practice. This robustness benefits from simultaneous consideration SAR image formation and oil spill segmentation, in which the intrinsic oil spill characteristics are modelled to guide the segmentation towards the oil spill areas in SAR images. Thus, the proposed method is capable of achieving robust and effective oil spill segmentation.

\subsection{Comparison with the Models with Different Distributions}
\label{comparison with distribution models}
\begin{figure*}	
\color{red}\begin{center}
\includegraphics[width=1.1\textwidth,height=0.66\textheight,center]{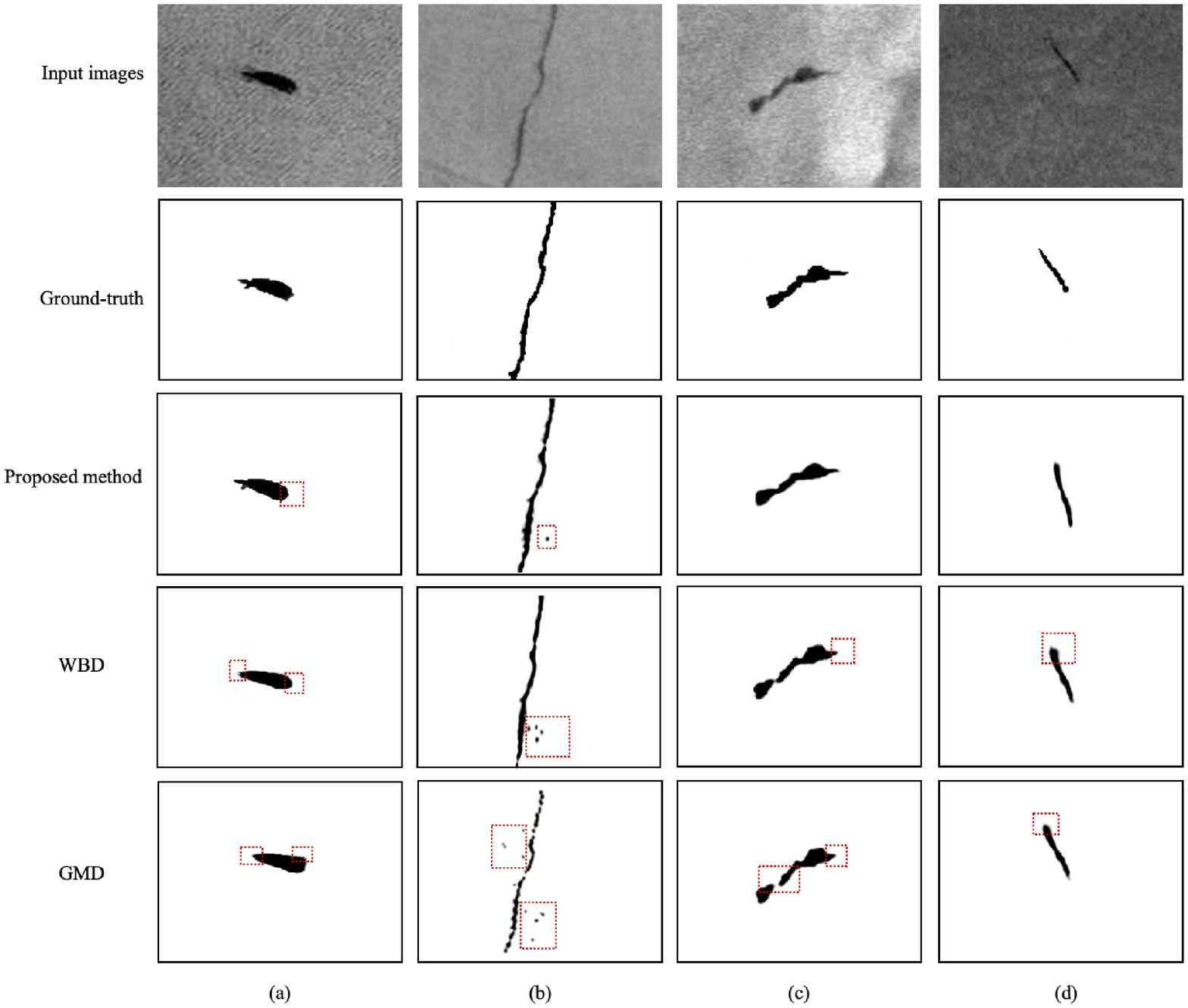}
\end{center}
\vspace*{-14mm}
\caption[]{The segmentation of oil spill images with the exploitation of the proposed method, the Weibull distribution (WBD) and the Gamma distribution (GMD) segmentation models. From top to bottom of columns (a)-(d) are the original oil spill images, the ground-truth, and the corresponding segmentation results of different segmentation models. The red dashed-line boxes are used to indicate the incorrect segmentations.}
\label{comparison with other distribution models}	
\end{figure*}

To evaluate the segmentation performance of the proposed method one step further, in this subsection, we compare the segmentation performance of the proposed method against that those with different distributions. Specifically, we compare the segmentation performance of the proposed method against the models with Weibull (WBD) and Gamma (GMD) distributions which are two representatives in remote sensing \cite{oliver2004understanding} \cite{gao2019characterization}, and their segmentation comparisons are shown in Fig. \ref{comparison with other distribution models}. Specifically, in this figure, from top to bottom in each column include the original oil spill SAR image, the ground-truth segmentation and the segmentations by the proposed method, the WBD and the GMD models, respectively. Examining these segmentation results, it is clear that the segmentations of our proposed method are much closer to the ground-truth segmentations, compared with those of the WBD and the GMD models for oil spill image segmentation.

Additionally, along with the qualitative experimental results, we compare the segmentation of different segmentation models in terms of their statistical segmentation accuracy over one-hundred oil spill images, shown in Table \ref{Accuracy of oil spill image segmentation with different segmentation models}. From the Table, it is observed that our proposed segmentation method achieves higher accuracy mean value with smaller standard deviation than the WBD and the GMD models. This demonstrates that our proposed method operates more accurate and stable segmentation. Thus, these evaluations from both qualitative and statistical quantitative perspectives have validated the effectiveness of our proposed method.

\begin{table}[h]
  \renewcommand\arraystretch{2.3}
	\centering
	\tabcolsep 0.05in
	\caption{SEGMENTATION PERFORMANCE IN TERMS OF STATISTICAL ACCURACY.}
	\begin{tabular}{c|c|ccc}
		\hline
		\hline
		\multicolumn{2}{c|}{\diagbox[width=4.0cm,dir=NW]{Image}{ Statistical Accuracy}{Method}} & Proposed Method & WBD & GMD \\
		\hline
		\multirow{2}{1.2cm}{ERS SAR}
		&mean&  $0.9916$ & $0.9847$ & $0.9618$ \\
		\cline{2-5}
		&standard deviation&  $0.0002$ & $0.0003$ & $0.0004$ \\
		\hline
		\multirow{2}{1.2cm}{Envisat ASAR }
		&mean&  $0.9917$ & $0.9858$ & $0.9644$\\
		\cline{2-5}
		&standard deviation& $0.0002$ & $0.0003$ & $0.0003$ \\
		\hline
		\hline
	\end{tabular}	
	\label{Accuracy of oil spill image segmentation with different segmentation models}
\end{table}

\subsection{Comparison with Neural Network Technique}
\label{neural segmentation comparison method for all types images}
To validate the segmentation performance of our proposed method comprehensively, in this subsection, we compare the segmentation performance of our proposed method with that of the standard U-Net neural network. The U-Net segmentation technique was developed to perform image segmentation without requiring a large dataset for training, fitting to our oil spill SAR image segmentation research where a large amount of oil spill images is difficult to obtain.

\begin{figure*}[htbp]
  \centering
  \subfigure[]{
  \includegraphics[width=0.86\textwidth,height=0.268\textheight]{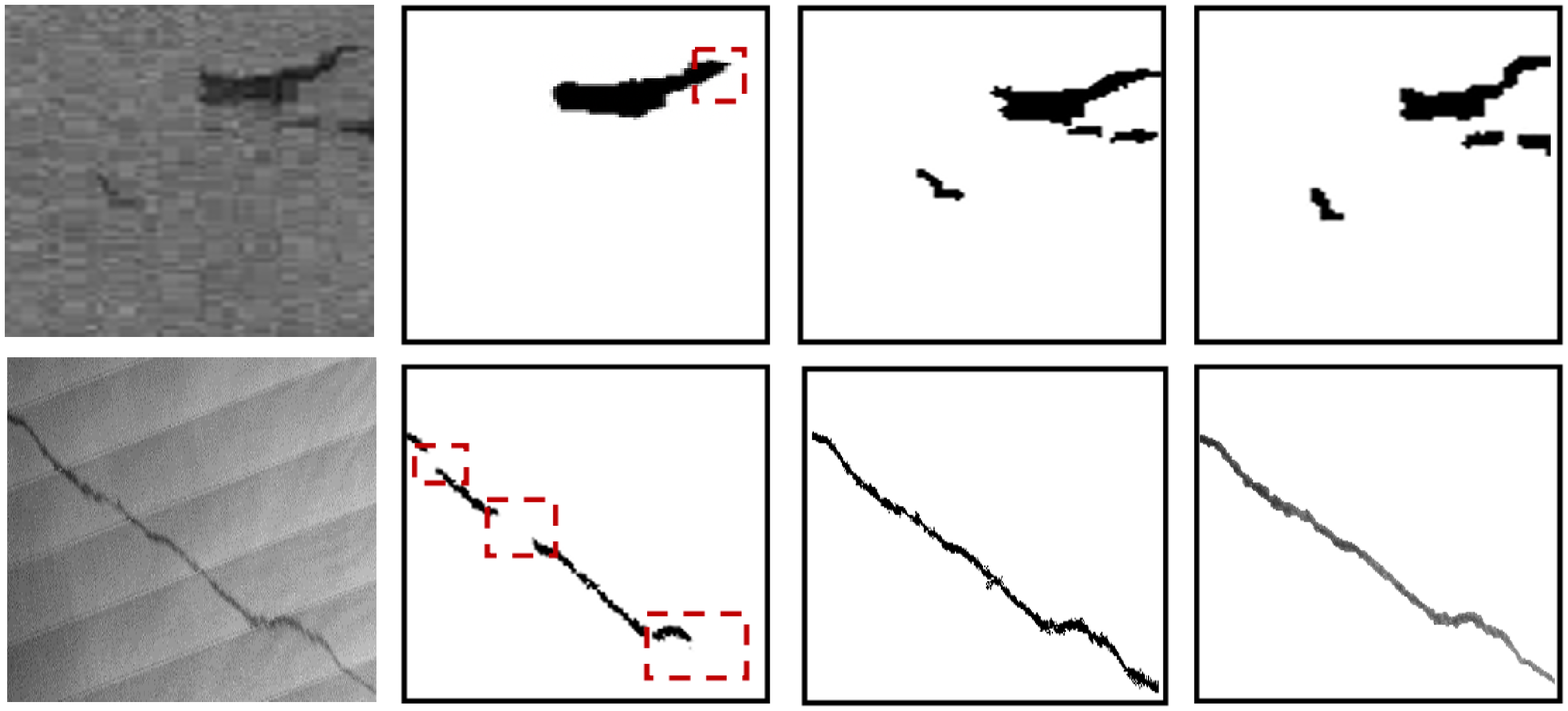}}
  \subfigure[]{
  \includegraphics[width=0.86\textwidth,height=0.268\textheight]{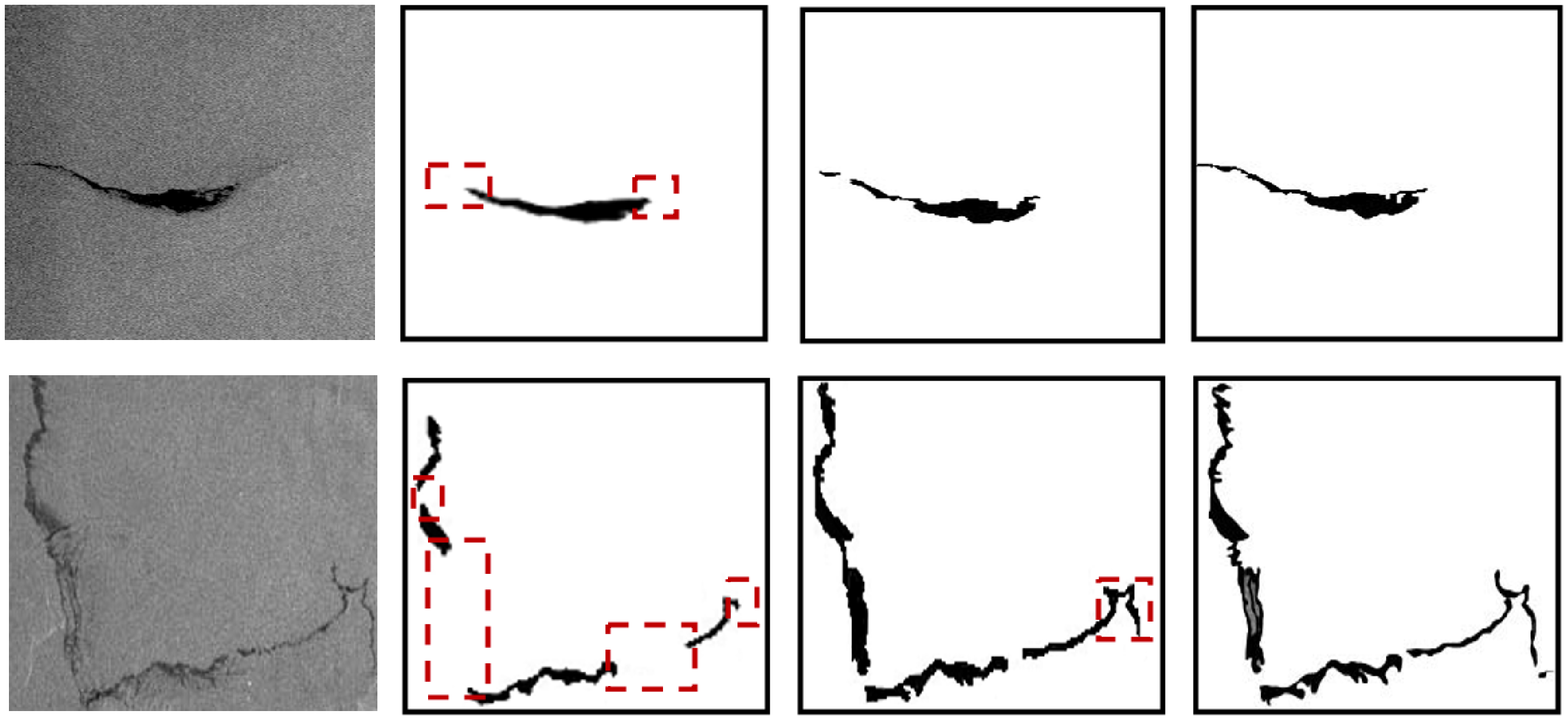}}
  \subfigure[]{
  \includegraphics[width=0.86\textwidth,height=0.268\textheight]{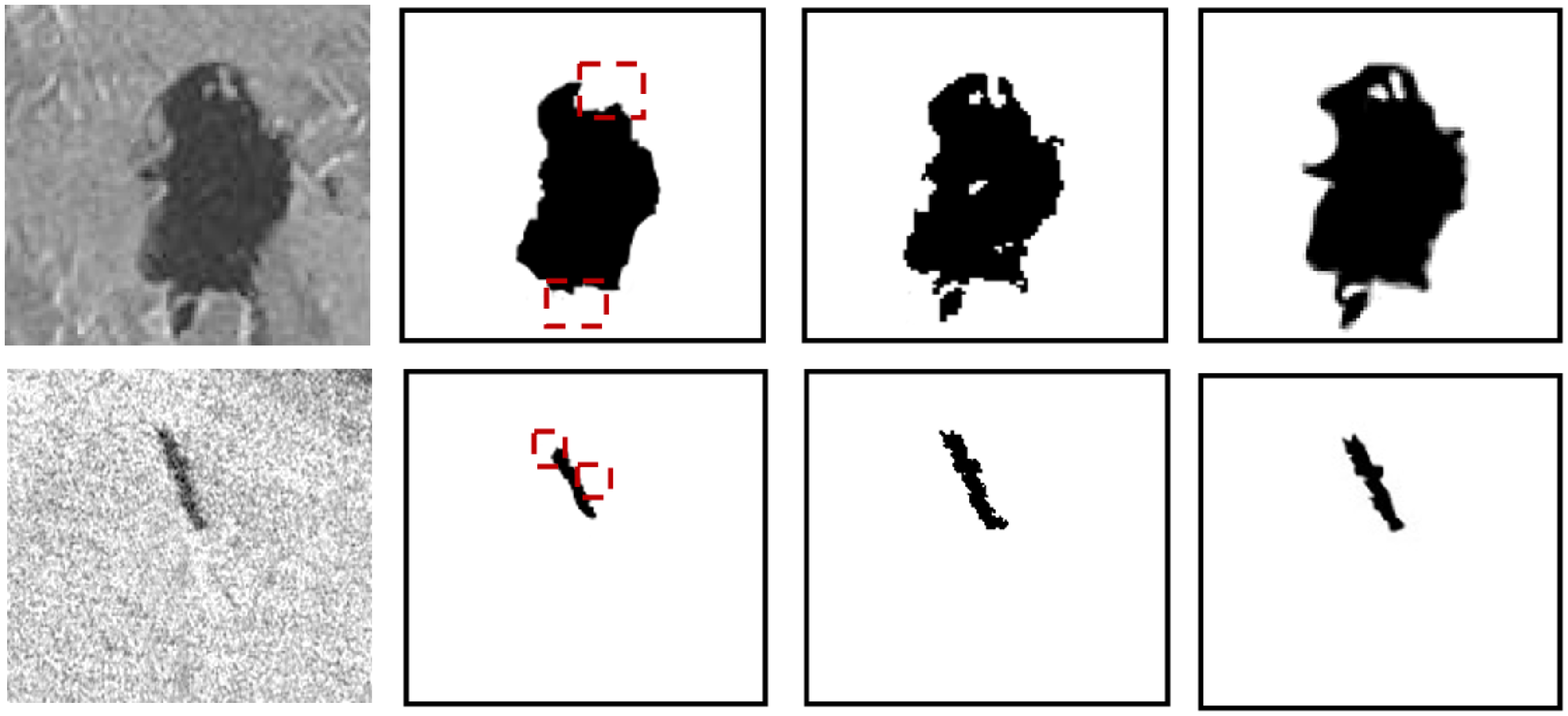}}
  \caption{Comparisons of oil spill segmentation between the U-Net neural network and our proposed method on ERS-1, ERS-2 and Envisat oil spill images. Specifically, (a), (b) and (c) illustrate the segmentation operation on ERS-1, ERS-2 and Envisat oil spill images respectively. In this figure, from left to right are the original oil spill images, oil spill segmentation with the U-Net neural network and our proposed method, and ground truth separately. The red dashed boxes show the incorrect segmentation areas.}
  \label{comparison with neural network for multi images}
\end{figure*}

We evaluate the segmentation performance of our proposed method and the U-Net neural network \cite{ronneberger2015u} with ERS-1, ERS-2 and Envisat oil spill images with different oil spill areas, and the segmentation results are shown in Fig. \ref{comparison with neural network for multi images}. Specifically, Fig. \ref{comparison with neural network for multi images} (a), (b) and (c) illustrate the segmentation on ERS-1, ERS-2 and Envisat oil spill images, respectively. In these subfigures, from left to right are the oil spill images, the segmentation results of the U-Net architecture, our proposed method segmentation and the ground truth. The segmentation results show that our method achieves more accurate oil spill segmentation. In addition, we compare the segmentation accuracy of different methods shown in Table \ref{Accuracy with neural network}. This Table shows that our method outperforms the U-Net architecture for oil spill image segmentation.

\begin{figure*}[htbp]
	\begin{minipage}[t]{2.3in}
    	\centering
    	\includegraphics[width=1.1\textwidth,height=0.23\textheight]{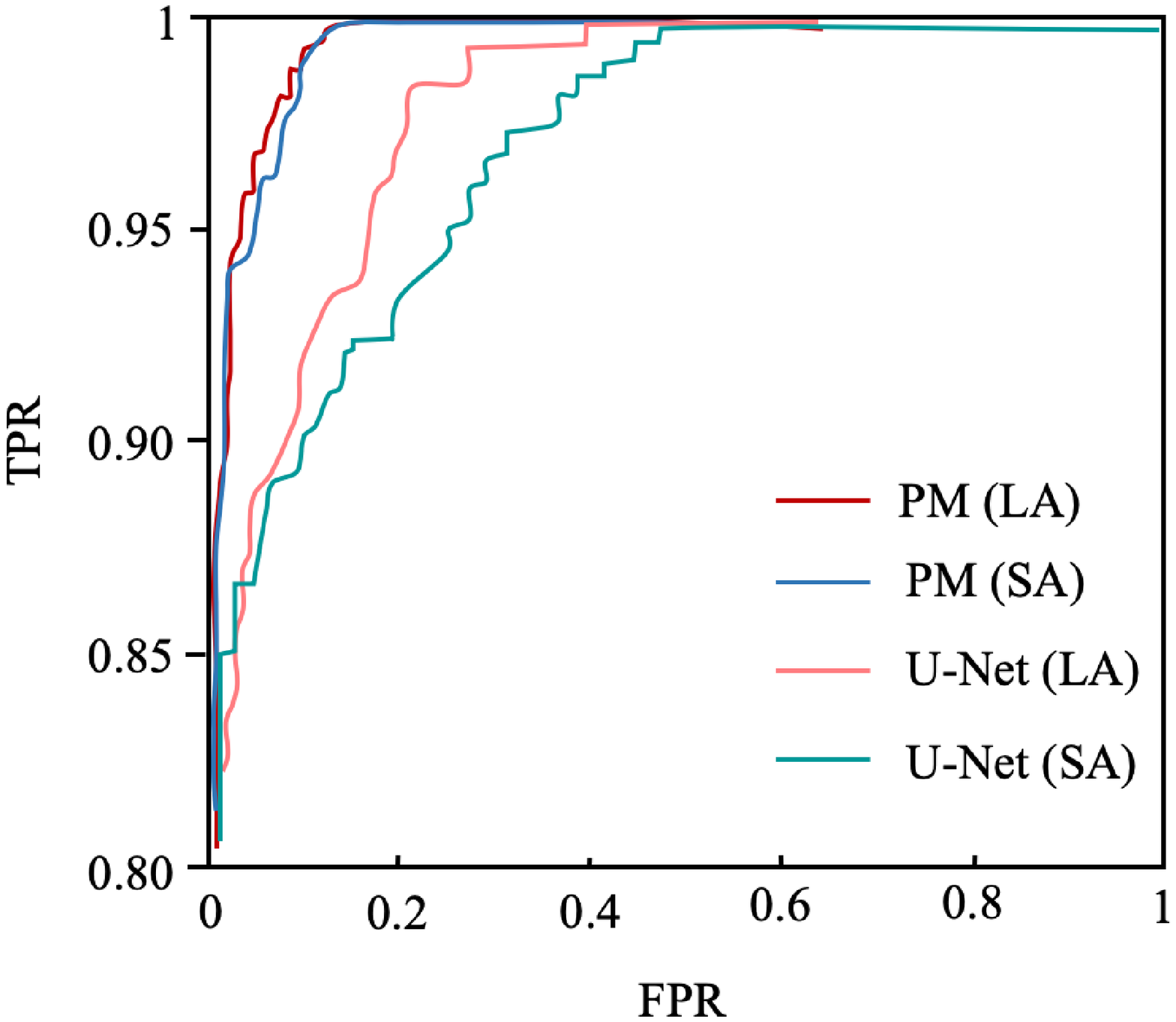}
    	\caption*{(a)}
	\end{minipage}
	\begin{minipage}[t]{2.3in}
    	\centering
    	\includegraphics[width=1.1\textwidth,height=0.23\textheight]{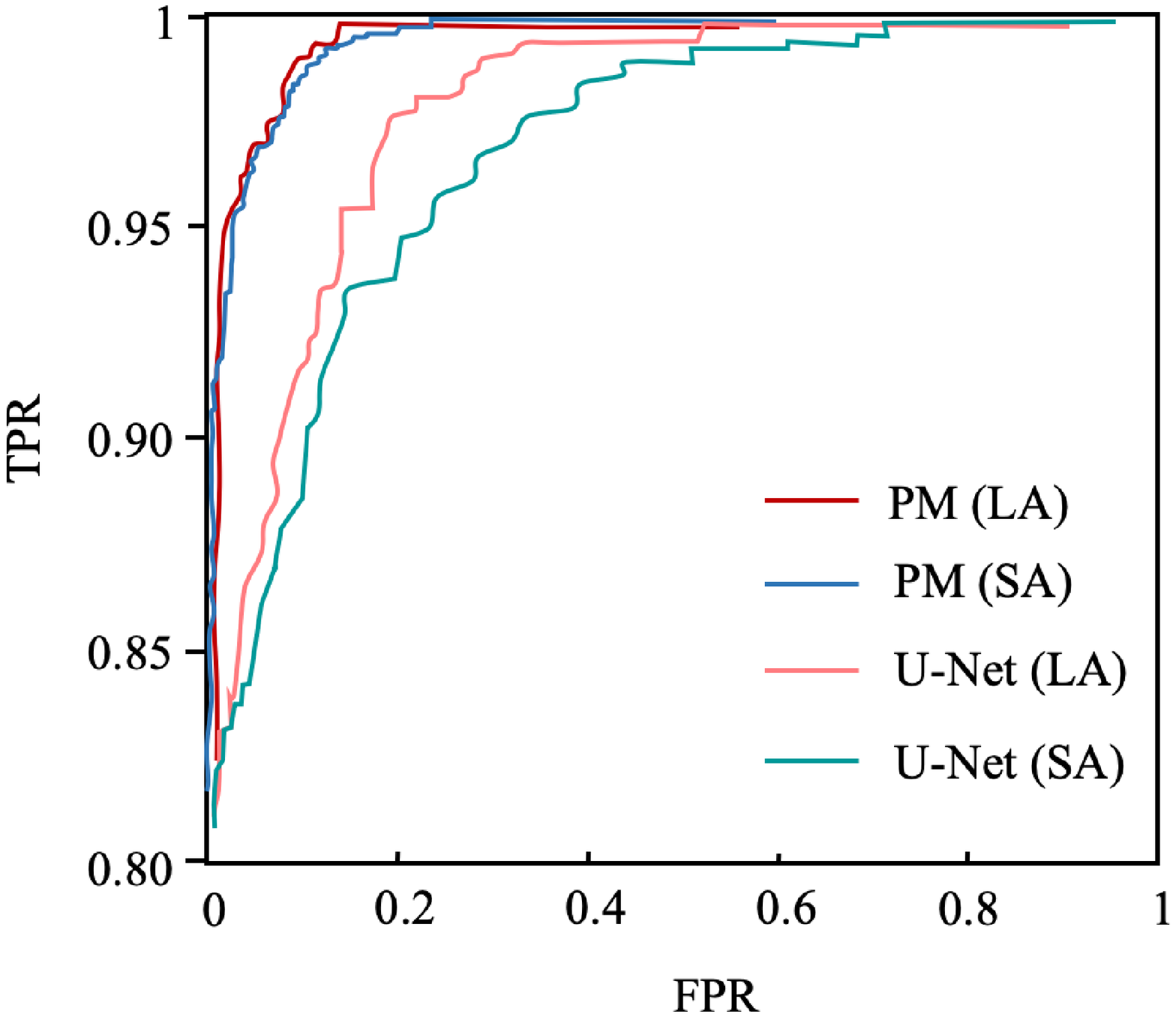}
     	\caption*{(b)}
	\end{minipage}
	\begin{minipage}[t]{2.3in}
    	\centering
    	\includegraphics[width=1.1\textwidth,height=0.23\textheight]{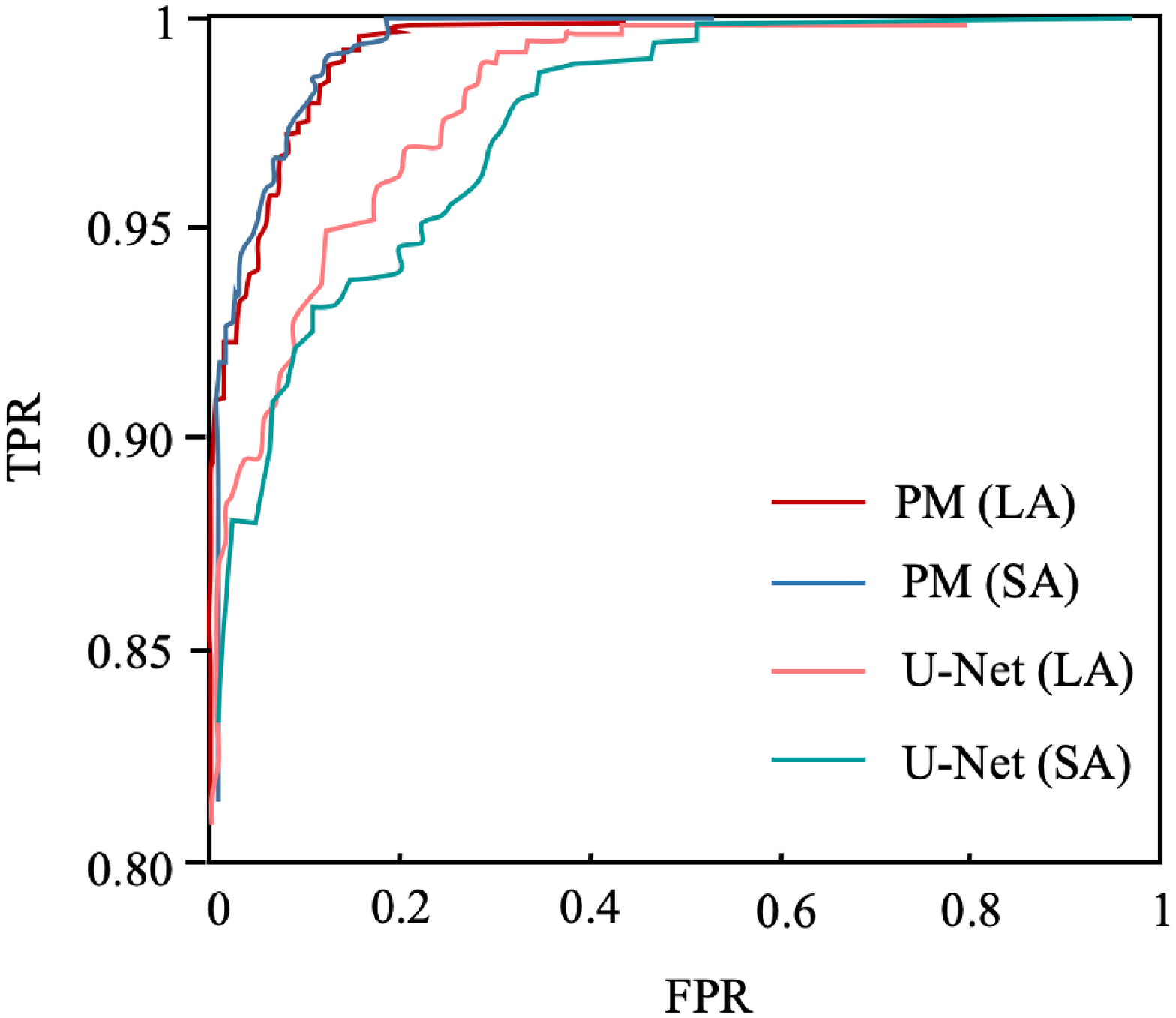}
     	\caption*{(c)}
	\end{minipage}
	\caption[]{ROC curves show the true positive rate (TPR) against the false positive rate (FPR) for oil spill segmentation outcomes. From left to right are the segmentation results of ERS-1, ERS-2 and Envisat oil spill images respectively.}
	\label{ROC courve for oil spill segmentation}		
\end{figure*}

\begin{table}[h]
	\renewcommand\arraystretch{2.3}
	\centering
	\tabcolsep 0.18in
	\caption{OIL SPILL SEGMENTATION ACCURACY BY U-Net AND OUR METHOD.}
	\begin{tabular}{c|c|cc}
		\hline
		\hline
		\multicolumn{2}{c|}{\diagbox[width=4.0cm,dir=NW]{Image}{ Accuracy}{Method}} & U-Net &  Proposed Method  \\
		\hline
		\multirow{2}{1.6cm}{ERS-1}
		&1&  $0.7629$ & $0.9469$  \\
		\cline{2-4}
		&2&  $0.7983$ & $0.9385$ \\
		\hline
		\multirow{2}{1.6cm}{ERS-2 }
		&1&  $0.8492$ & $0.9286$ \\
		\cline{2-4}
		&2& $0.7921$ & $0.9261$  \\
		\hline
		\multirow{2}{1.6cm}{Envisat}
		&1& $0.8369$ & $0.9582$  \\
		\cline{2-4}
		&2& $0.8123$ & $0.9507$ \\
		\hline
		\hline
	\end{tabular}	
	\label{Accuracy with neural network}
\end{table}

From Fig. \ref{comparison with neural network for multi images} and Table \ref{Accuracy with neural network}, we observe that the segmentation performance of the U-net technique depends on the size of oil spill areas in SAR images. Thus, we conduct further evaluation for the segmentation results shown in Fig \ref{ROC courve for oil spill segmentation}. Specifically, in Fig \ref{ROC courve for oil spill segmentation}, PM (LA) and PM (SA) refer to the segmentations of our proposed method for SAR images with larger and slender oil spill areas respectively, and the other two stand for the U-Net neural net segmentation for the individual images. From this figure, we have that for both larger and slender oil spill area segmentation, our method achieves better oil spill segmentation than the U-Net architecture.
Evaluations from both visual and quantitative results validate that our proposed method outperforms the U-Net architecture for oil spill image segmentation.

\begin{figure*}	
\begin{center}
\includegraphics[width=1.1\textwidth,height=0.64\textheight,center]{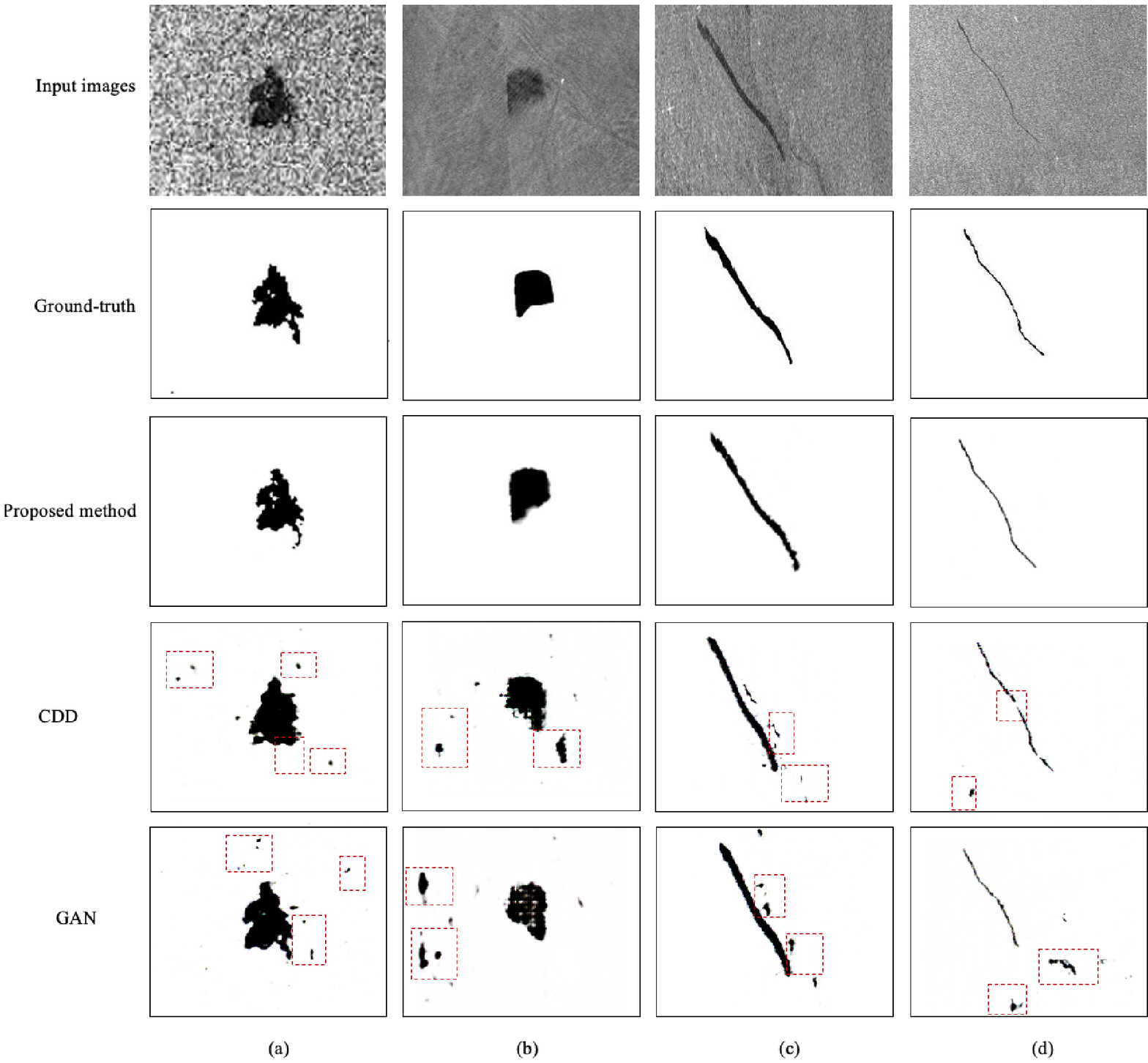}
\end{center}
\vspace*{-14mm}
\caption[]{The segmentation of oil spill images with the exploitation of the proposed method, the CDD and the GAN neural net techniques. From top to bottom of columns (a)-(d) are the original oil spill SAR images, the ground-truth, and the corresponding segmentation results of different segmentation methods. The red dashed-line boxes are used to indicate the incorrect segmentation areas.}
\label{comparison with other neural networks}	
\end{figure*}

In addition to more comprehensive validation for our proposed method, we compare the segmentation of the proposed method against the capacitory discrimination distance (CDD) divergence learning neural net \cite{yu2018oil} and the generative adversarial network (GAN) \cite{goodfellow2014generative}. Specifically, the CDD technique is regarded as a segmentation method which is able to generate segmentation maps for irregular oil spill areas in an accurate way. GAN operates the segmentation by adversarial training two models to play against each other. This increases the model representational power for oil spill image segmentation. In this scenario, these two techniques are more applicable for practical oil spill image segmentation and thus we choose them as comparison methods, and the segmentation results are shown in Fig. \ref{comparison with other neural networks}. From this figure, it is observed that our proposed segmentation method achieves better performance than CDD and GAN for oil spill image segmentation.

\subsection{Discussions}
In this subsection, we discuss the performance of the proposed method in the presence of low backscattering areas and its convergence analysis.
\subsubsection{Understanding the Behavior of the Proposed Method in the Presence of Low Backscattering Areas}
For the backscattering from the low backscattering areas that is sufficiently different from the oil slicks, our proposed segmentation method is able to achieve accurate oil spill segmentation. However, in the condition that the oil slick area and the low backscattering region look similar, the segmentation performance varies case by case. Specifically, in our experimental validation with different types of oil spill SAR images, we found that if the low backscattering region is away from the oil spill area in the SAR images, the segmentation performance is not degenerated. In contrast, if the low backscattering region is close to the oil spill area, the segmentation performance is slightly affected.

To implement oil spill image segmentation, we initialise the segmentation process by determining the starting contour. Traditionally, the initialisation region is does not contain low backscattering areas. Thus, in the operation of segmentation, active contour can move towards the oil spill areas using our proposed method. However, if low backscattering areas fall within the initialisation zone, this causes interference to oil spill image segmentation. For small or tiny low backscattering areas falling within the starting contour, our proposed segmentation method is able to handle this challenge. For large low backscattering areas, we will address the challenge in the future research.

\subsubsection{Theoretical Explanation for the Convergence of the Proposed Method}
In this part, we analyse the proposed method for its segmentation convergence. Particularly, we will analyse the monotonicity and boundedness of the segmentation energy functional $E_{S}$ below.

As introduced in Section \ref{oil spill segmentation formulation}, the proposed method consists of two types of terms: the level set fitting energy term and the regularisation terms. The level set fitting energy term $E_{F}$, which plays a dominate role in the evolution of minimising $E_{S}$, is formulated in Eq.(\ref{level set energy functional for fitness}), and the gradient for iteratively optimising the fitting energy functional $E_{F}$ is: $\frac{\partial E_{F}}{\partial \phi}=\big(\gamma_{2}\breve{\varepsilon}_{2}-\gamma_{1}\breve{\varepsilon}_{1}\big)\delta_{\epsilon}\big(\phi\big)$, where $\breve{\varepsilon}_{1}$ and $\breve{\varepsilon}_{2}$ can be obtained from Eq. (\ref{intermidiate term}). We then investigate the convergence of the objective energy functional $E_{S}$ in terms of monotonicity and boundedness, related to the gradient $\frac{\partial E_{F}}{\partial \phi}=\big(\gamma_{2}\breve{\varepsilon}_{2}-\gamma_{1}\breve{\varepsilon}_{1}\big)\delta_{\epsilon}\big(\phi\big)$. In the gradient $\frac{\partial E_{F}}{\partial \phi}=\big(\gamma_{2}\breve{\varepsilon}_{2}-\gamma_{1}\breve{\varepsilon}_{1}\big)\delta_{\epsilon}\big(\phi\big)$, $\breve{\varepsilon}_{1}$ and $\breve{\varepsilon}_{2}$ are the oil spill fitting and the background fitting differences, and $\gamma_{1}$ and $\gamma_{2}$ are positive weights that balance the oil spill region and background fitting, respectively. The positive weights are empirically set subject to $\gamma_{1}<\gamma_{2}$. There are two reasons for this setting-up. First, in practice, the detected oil spill edges normally evolve from an arbitrary contour loosely surrounding the oil spill regions to the correct oil spill regions, such that the major evolutions take place outside the oil spill regions. Secondly, according to \cite{ulaby2014microwave}, the background outside the oil spill regions is less homogeneous than the oil spill regions. The background fitting difference $\breve{\varepsilon}_{2}$ requires a heavy weight for encouraging accurate fitting. We thus practically set the oil spill region fitting weight $\gamma_{1}$ to be smaller than the background fitting weight $\gamma_{2}$. Additionally, as background is less homogeneous than the oil spill regions, the background fitting difference $\breve{\varepsilon}_{2}$ is generally greater than the oil spill fitting difference $\breve{\varepsilon}_{1}$. Therefore, the gradient keeps positive in the evolution, reflecting a monotonous increase during the evolution. On the other hand, the boundedness is enabled by the regularisation terms. Specifically, the regularisation terms shown in Eqs. (\ref{contour regularization term}) and (\ref{regularization term}) constrain the oil contour length and update regularity, respectively. They thus prevent the evolution from infinity and form a bound to the optimisation of the segmentation energy functional.

\section{Conclusions}
We have presented our novel oil spill SAR image segmentation method by simultaneously considering oil spil SAR image formation and oil spill segmentation. Specifically, we commenced by exploring SAR imaging for marine oil spill observation to obtain the probability distribution representation of SAR images. The probability distribution was modelled with the intrinsic oil spill characteristics which distinguish the oil spills from the background. We then exploited the probability distribution representation to construct the segmentation energy functional to implement oil spill segmentation. The incorporated probability distribution representation enhances the representational power of the segmentation energy functional. Particularly, in our proposed segmentation method, benefiting from the incorporation of the oil spill SAR image formation, the intrinsic characteristics of oil spills are exploited in favor of guiding the segmentation to operate intentionally towards the oil spill areas to perform accurate oil spill segmentation. Experimental results have shown the effectiveness of our proposed segmentation method, compared against several state-of-the-art segmentation methodologies.


\ifCLASSOPTIONcaptionsoff
  \newpage
\fi

\bibliographystyle{IEEEtran}
\bibliography{reference}

\section*{Acknowledgments}
Fang Chen is supported by China Scholarship Council (No. 201806450015), and the PhD studentship from University of Leicester. Dr. A. Zhang acknowledges the Start-Up Research Fund support from UIC under Grant UICR0700015-22.

\begin{IEEEbiography}[{\includegraphics[width=1in,height=1.25in,clip,keepaspectratio]{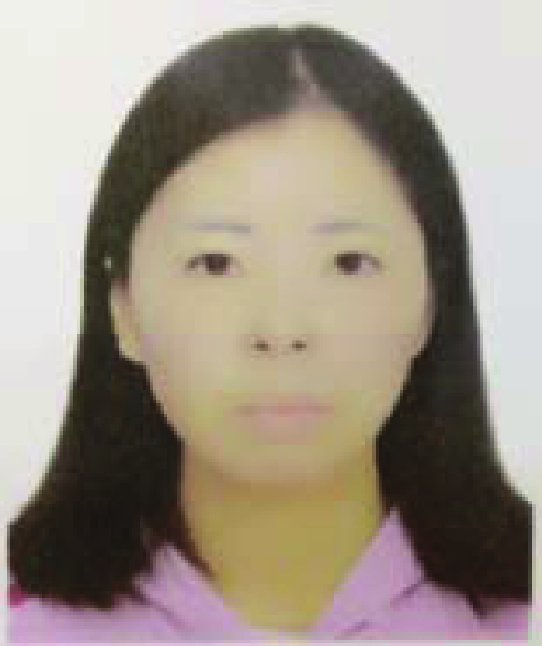}}]{Fang Chen}
received the B. Eng. degree in communication engineering from Southwest University for Nationalities, Chengdu, China, and the M. Eng. degree in information and communication engineering from China University of Petroleum (East China), Qingdao, China. She is currently pursuing the Ph. D. degree at School of Computing and Mathematical Sciences, University of Leicester, Leicester, U.K.
	
Her current research interests include machine learning and image processing, with application to marine remote sensing synthetic aperture radar image classification and segmentation.
\end{IEEEbiography}

\begin{IEEEbiography}[{\includegraphics[width=1in,height=1.25in,clip,keepaspectratio]{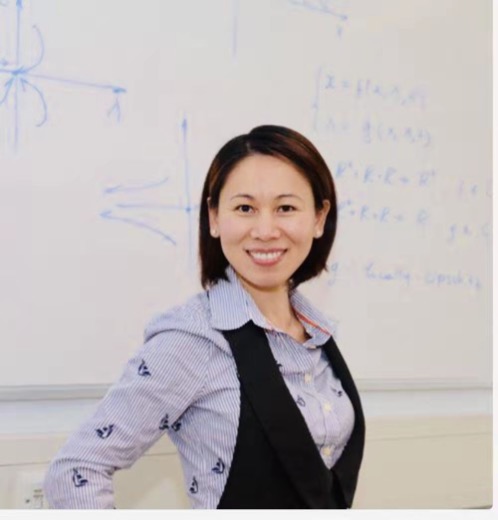}}]{Dr. A. Zhang }
is currently an Associate Professor in Statistics at Beijing Normal University – Hongkong Baptist University UIC. She has a strong inter-disciplinary background in Applied Statistics, Data Analytics, Financial Mathematics, Actuarial Science and Bayesian Statistics. Dr. Zhang is the author of the paper entitled “New Findings on Key Factors Influencing the UK’s Referendum on Leaving the EU”, published in the World Development journal, which was widely reported by both the UK (including BBC, The Huffington Post and The Independent) and other international media (such as Yahoo News, Insider Higher Ed), including interviews with The Conversation (Zhang, 2017) and Radio Sputnik World Service (Moscow).
\end{IEEEbiography}

\begin{IEEEbiography}[{\includegraphics[width=1in,height=1.25in,clip,keepaspectratio]{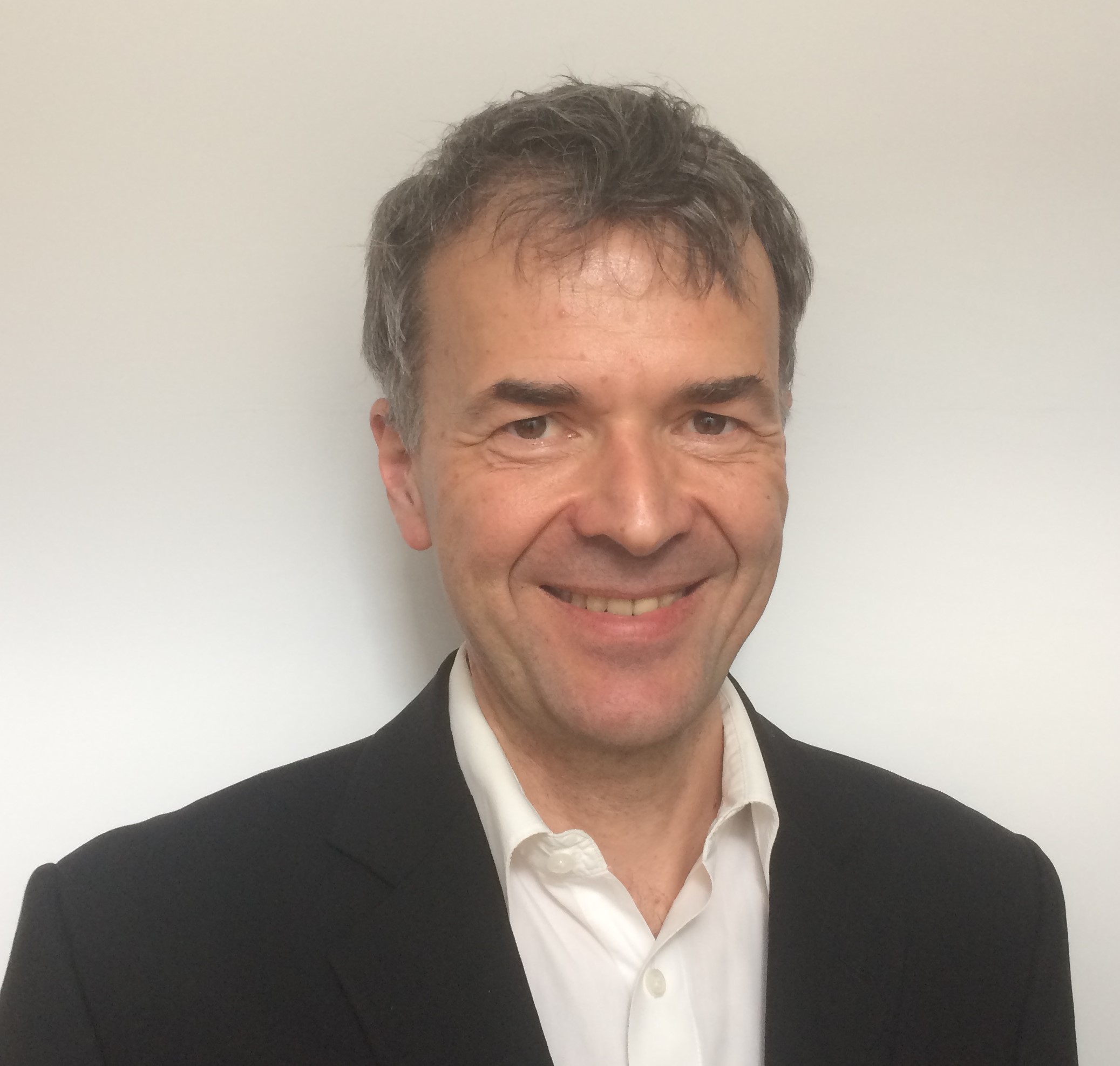}}]{Heiko Balzter}
is Professor of Physical Geography at the University of Leicester. He leads the UKRI Strategic Programme Coordination Team on “Landscape Decisions” (£10.5 million) and is a member of the National Centre for Earth Observation. He holds the Royal Society’s Wolfson Research Merit Award (2011), Royal Geographical Society’s Cuthbert Peek Award ‘for advancing geographical knowledge of human impact through earth observation’ (2015) and Copernicus Masters ‘Sustainable Living Award’ (2017) for his work on deforestation monitoring. His 120 journal papers have been cited $>$6000 times (h-index=42; i10-index=98, Google Scholar).
Prof. Balzter is alternate UK representative on the GEO Programme Board and is leading the Copernicus Land Monitoring Service for the UK. He is involved in collaborative research funded by the European Space Agency, the UK Space Agency and the Natural Environment Research Council. His research interests include EO of the land surface and spatial-temporal patterns and processes. His forest research includes work on above-ground biomass, logging detection in near-real-time, fire monitoring and tree disease detection.
Before joining Leicester, he worked at the Centre for Ecology and Hydrology (1998-2006). He was Head of Department of Geography in Leicester from 2008-2011. He graduated with a PhD from Justus-Liebig-University, Giessen, Germany, in 1998.
\end{IEEEbiography}

\begin{IEEEbiography}[{\includegraphics[width=1in,height=1.25in,clip,keepaspectratio]{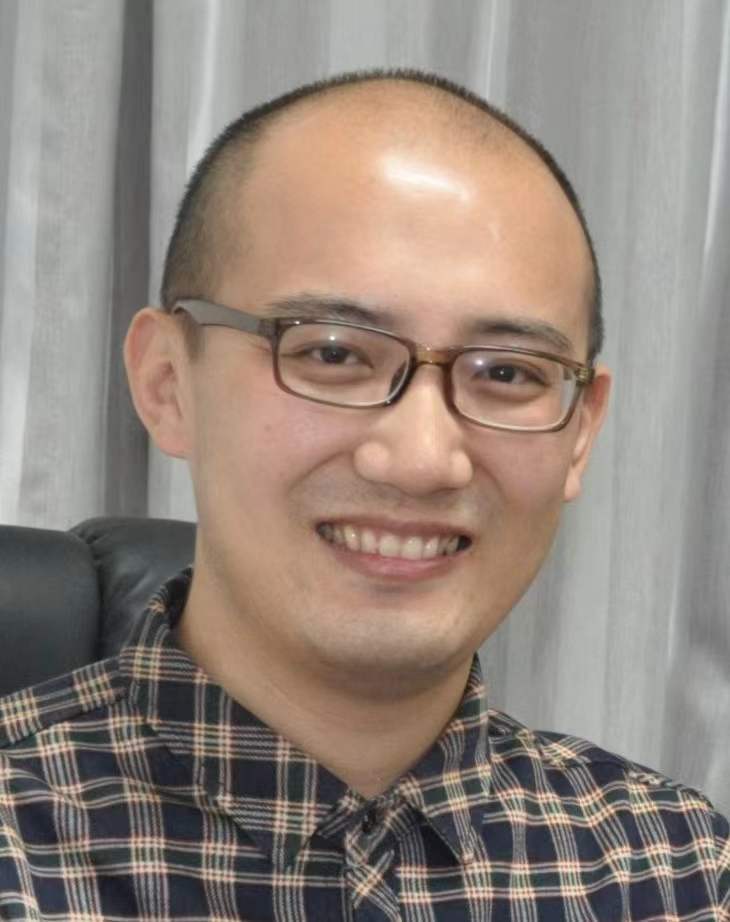}}]{Peng Ren}
(Senior Member, IEEE) received the B.Eng. and M.Eng. degrees in electronic engineering from the Harbin Institute of Technology, Harbin, China, and the
Ph.D. degree in computer science from the University of York, York, U.K.
He is currently a Professor with the College of Oceanography and Space Informatics, China University of Petroleum (East China), Qingdao, China. His research interests include marine information processing, remote sensing,  machine learning, etc. Dr. Ren was a recipient of the K. M. Scott Prize from the University of York in 2011 and the Eduardo Caianiello Best Student Paper Award at the 18th International Conference on Image Analysis and Processing in 2015 as a coauthor.
\end{IEEEbiography}

\begin{IEEEbiography}[{\includegraphics[width=1in,height=1.25in,clip,keepaspectratio]{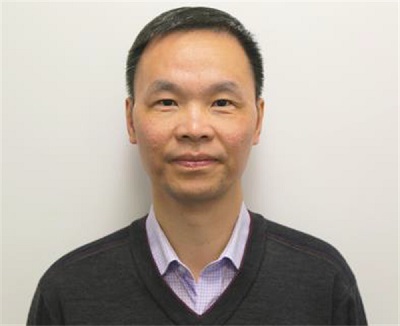}}]{Prof. Huiyu Zhou}
received a Bachelor of Engineering degree in Radio Technology from Huazhong University of Science and Technology of China in 1990, and a Master of Science degree in Biomedical Engineering from University of Dundee of United Kingdom in 2002, respectively. He was awarded a Doctor of Philosophy degree in Computer Vision from Heriot-Watt University, Edinburgh, United Kingdom, 2006. Dr. Zhou currently is a full Professor at School of Computing and Mathematical Sciences, University of Leicester, United Kingdom. He has published over 380 peer-reviewed papers in the field.
\end{IEEEbiography}


\end{document}